\documentclass[10pt,twocolumn,letterpaper]{article}

\usepackage{cvpr}              % To produce the CAMERA-READY version
\usepackage{amssymb}
\usepackage{bbding}
\usepackage{pifont}
\usepackage{wasysym}
\usepackage{utfsym}
\usepackage{fontawesome}
\usepackage{multirow}
\usepackage[table]{xcolor}
\definecolor{cvprblue}{rgb}{0.21,0.49,0.74}
\usepackage[pagebackref,breaklinks,colorlinks,citecolor=cvprblue]{hyperref}
\usepackage{makecell}
\usepackage{bbding}
\usepackage{float}
\usepackage{color}
\usepackage{enumitem}
\usepackage{utfsym}
\usepackage{xspace}
\usepackage{pifont}
\usepackage{arydshln}
\usepackage[T1]{fontenc}
\def\paperID{13066} % *** Enter the Paper ID here
\def\confName{CVPR}
\def\confYear{2025}

\usepackage{sidecap}
\usepackage{booktabs}

\usepackage{afterpage}

\title{MIMO: A medical vision language model with visual referring multimodal input and pixel grounding multimodal output}
\author{%
  Yanyuan Chen$^{1}$\textsuperscript{$\dagger$},~~Dexuan Xu$^{2}$\textsuperscript{$\dagger$},~~Yu Huang$^{3}$\textsuperscript{\textnormal{*}}, ~~Songkun Zhan$^{1}$,~~Hanpin Wang$^{2}$ \\~~Dongxue Chen$^{4}$,  Xueping Wang$^{4}$,~~Meikang Qiu$^{5}$,~~Hang Li$^{6}$\textsuperscript{\textnormal{*}}\\[0.25cm]
 {\fontsize{10.5pt}{12pt}\selectfont $^{1}$School of Software \& Microelectronics, Peking University, $^{2}$School of Computer Science, Peking University}\\
 {\fontsize{10.5pt}{12pt}\selectfont $^{3}$National Engineering Research Center for Software Engineering, Peking University}\\
 {\fontsize{10.5pt}{12pt}\selectfont $^{4}$Peking University Sixth Hospital, 
  $^{5}$ Augusta University, 
  $^{6}$ Peking University First Hospital}
}

\begin{document}
\maketitle
\thispagestyle{empty}

\begin{abstract}

Currently, medical vision language models are widely used in medical vision question answering tasks. However, existing models are confronted with two issues: for input, the model only relies on text instructions and lacks direct understanding of visual clues in the image; for output, the model only gives text answers and lacks connection with key areas in the image. To address these issues, we propose a unified medical vision language model \textbf{MIMO}, with visual referring \textbf{M}ultimodal \textbf{I}nput and pixel grounding \textbf{M}ultimodal \textbf{O}utput. MIMO can not only combine visual clues and textual instructions to understand complex medical images and semantics, but can also ground medical terminologies in textual output within the image. To overcome the scarcity of relevant data in the medical field, we propose \textbf{MIMOSeg}, a comprehensive medical multimodal dataset including 895K samples. MIMOSeg is constructed from four different perspectives, covering basic instruction following and complex question answering with multimodal input and multimodal output. We conduct experiments on several downstream medical multimodal tasks. Extensive experimental results verify that MIMO can uniquely combine visual referring and pixel grounding capabilities, which are not available in previous models. Our project can be found in \url{https://github.com/pkusixspace/MIMO}.
\renewcommand{\thefootnote}{$\dagger$}
\footnotetext{Equally contributing first authors.}
\renewcommand{\thefootnote}{\arabic{footnote}} 
\renewcommand{\thefootnote}{*}
\footnotetext{Yu Huang and Hang Li are corresponding authors.}
\renewcommand{\thefootnote}{\arabic{footnote}} 
\end{abstract}    
\section{Introduction}

\begin{figure}
  \centering
  \includegraphics[width=0.4\textwidth]{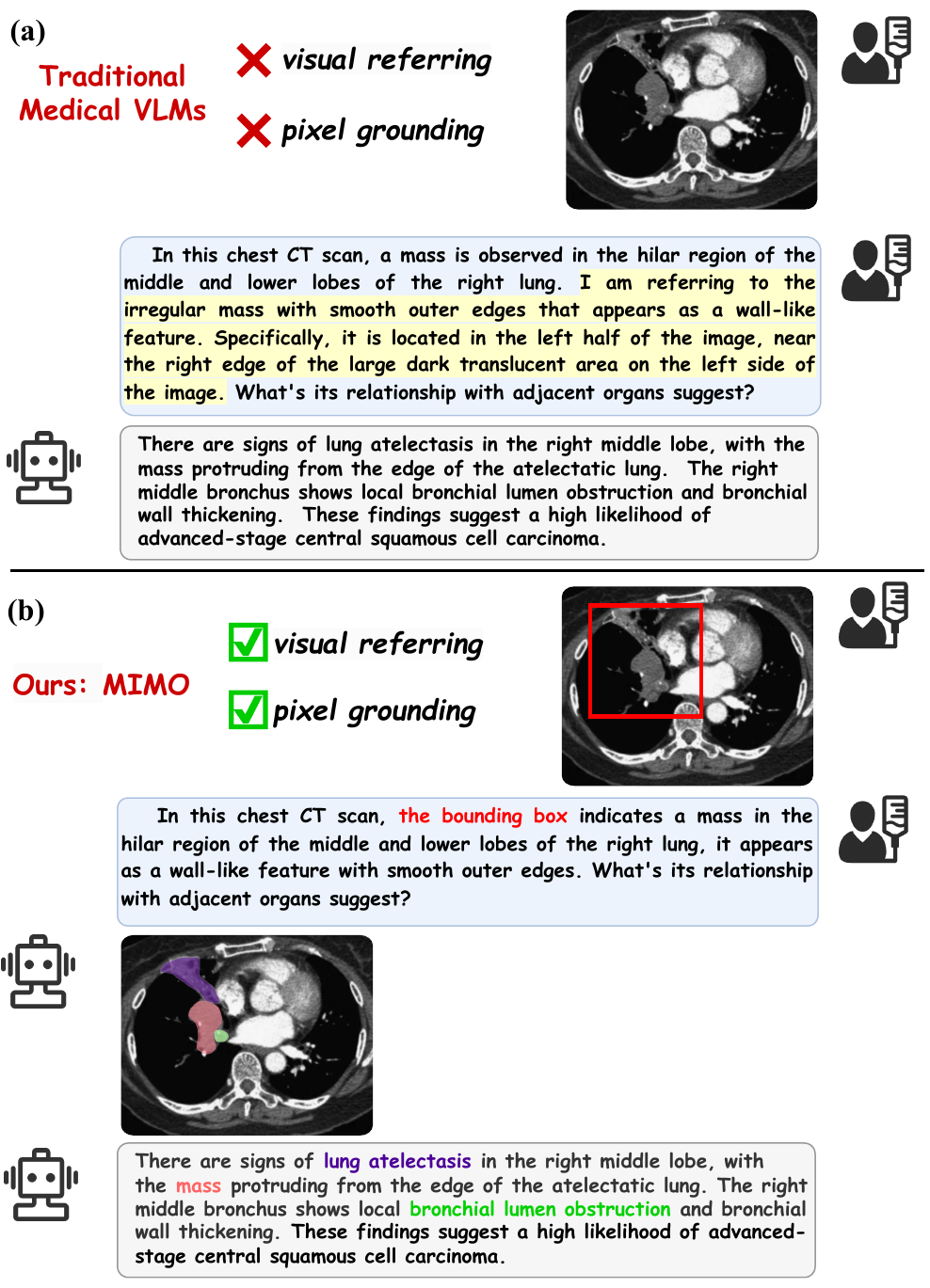} 
  \caption{An illustration of visual referring and pixel grounding capabilities.}
  \label{fig:intro}
  \vspace{-0.5cm}
\end{figure}

Currently,  medical vision-language models (MVLMs) equip large language models (LLMs) with visual perception modules for image understanding and leverage the powerful capabilities of LLMs to comprehend complex textual queries \cite{chen2024huatuogpt,li2023llava,moor2023med}, demonstrating outstanding performance in medical visual question answering (MVQA). However, current models mainly use textual instructions to query the entire image, and subsequently output textual responses for image-related question answering tasks. While this approach enables a rich variety of queries and responses on the text side, it exhibits significant shortcomings in MVLM: unlike general domains, most medical terminologies typically involve specialized expressions. The limitations of text-based input and output may lead to expressions that are either overly concise or excessively complicated. 

Figure \ref{fig:intro} provides an illustrative example. For questions related to specific areas of medical images, accurately describing the region of interest through language is challenging (shown in Figure \ref{fig:intro}(a)). Visual Question Answering, unlike unimodal question answering, requires close integration with specific image information. Although professional physicians can use anatomical terms (e.g., ``\textit{the hilar region of the middle and lower lobes of the right lung}'') to indicate these areas, however, in practical applications, the imaging findings of different patients may exhibit individual variations, further complicating the description. MVQA heavily relies on visual interpretability, even communication among specialized doctors is centered around image referring and grounding (shown in Figure \ref{fig:intro}(b)).

Inspired by the above gap, in this paper, we highlight the following two capabilities of the MVLM model: (i) visual referring in question inputs, and (ii)  pixel grounding in answering outputs. Visual referring enables the model to assist users in articulating questions by combining text and visual prompts to specify a particular region \cite{liu2023gres,you2023ferret,zellers2019recognition}, whereas pixel grounding requires the model to align semantic descriptions with segmentation masks within the image \cite{zhang2024groundhog,rasheed2024glamm,zhou2019grounded}. 

To construct the aforementioned MVLM system, two major challenges exist: (i) Lack of a unified model architecture. Due to design limitations, existing MVLM architectures \cite{chen2024huatuogpt,li2023llava,moor2023med} do not support multimodal input with visual prompts and multimodal output with segmentation masks. (ii) The scarcity of multimodal medical datasets that support these capabilities. Existing medical datasets \cite{xie2024medtrinity,chen2024huatuogpt,li2023llava} focus on either visual question answering or segmentation tasks, without referring and grounding annotations simultaneously.

Targeting these two challenges, we propose \textbf{MIMO}, a unified medical vision language model with visual referring \textbf{M}ultimodal \textbf{I}nput and pixel grounding \textbf{M}ultimodal \textbf{O}utput. First of all, MIMO models all visual prompts as embeddings in the same space as the image features. To effectively facilitate cross-modal alignment of these multimodal inputs, we propose a Multi-modal Input Aligner to bridge the modality gap. Then, MIMO leverages LLM as foundation, and decodes the segmentation tokens to obtain grounding masks, where these segmentation tokens are linked to semantic phrases in LLM' natural language output.  Equipped with above methods, MIMO can seamless integrate natural language responses with medical entity segmentations. To train MIMO, we propose a medical multi-modal referring and grounding dataset \textbf{MIMOSeg} with 895K samples. MIMOSeg is constructed from four different perspectives, covering basic instruction following and complex question answering with multimodal input and multimodal output. We conduct experiments on several downstream medical multimodal tasks. Extensive experimental results verify that MIMO can uniquely combine visual referring and pixel grounding capabilities, which are not available in previous models.  To the best of our knowledge, MIMO is the first work to simultaneously integrate visual referring and pixel grounding in medical vision language models.

Our contributions are as follows:

(1) We present \textbf{MIMO}, the first MVQA model that integrates visual referring multimodal input and pixel grounding multimodal output, specifically designed to address the unique challenges of medical image understanding and textual reasoning.

(2) We propose a comprehensive dataset \textbf{MIMOSeg} containing 895K samples, which is designed to facilitate the training of MIMO. This dataset provides a diverse range of scenarios, ensuring that the model can learn to handle both basic instructions and complex medical queries with different modalities of input.

(3) We conduct extensive held-in and held-out experiments to evaluate the performance of MIMO across various medical multimodal tasks. Experimental results verify that MIMO can uniquely combine visual referring and pixel grounding capabilities, which are not available in previous models.

\section{Related Work}
\label{sec:relatedwork}

\begin{figure*}
  \centering
  \includegraphics[width=1\textwidth]{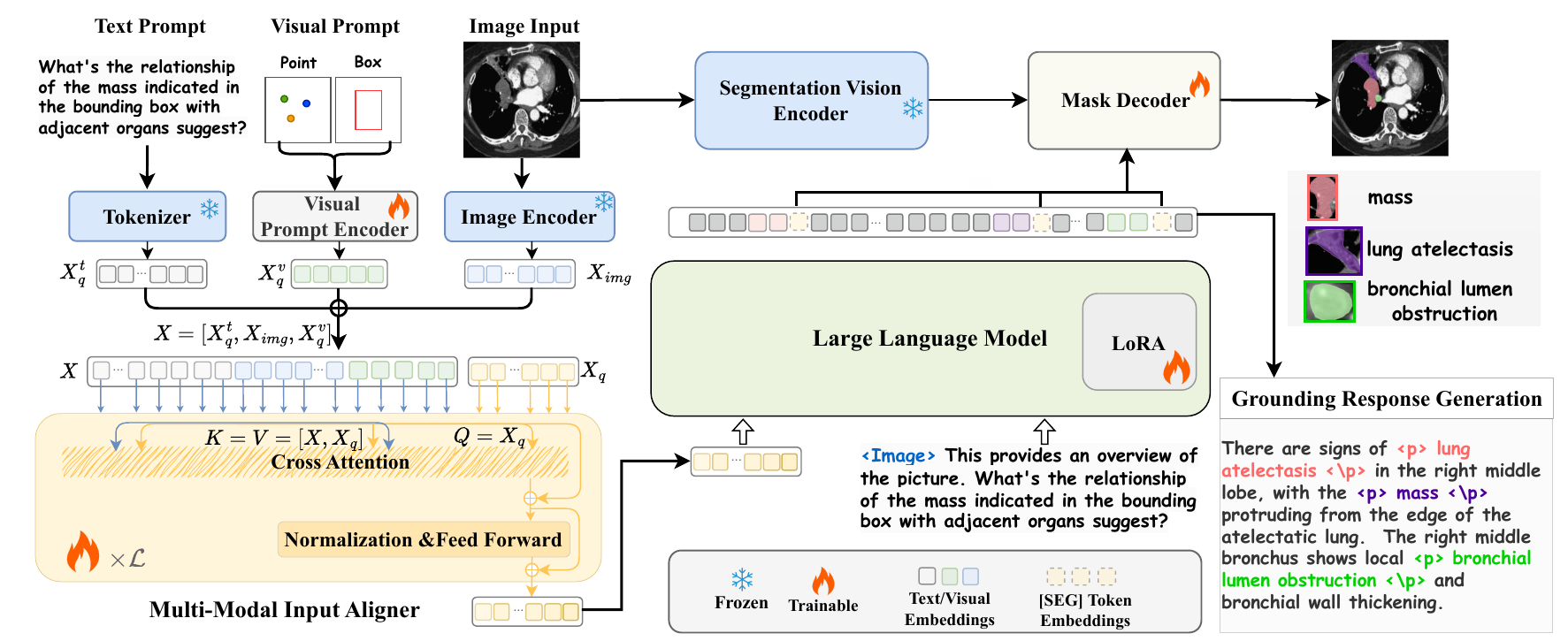} 
  \caption{The overall architecture of MIMO. The model consists of an image encoder, a visual prompt encoder, a LLM, a segmentation vision encoder and a mask decoder. While accepting optional visual input, the model can also provide visual segmentation results associated with the medical entities in the text response.}
  \label{fig:model}
  \vspace{-0.5cm}
\end{figure*}

\textbf{Multimodal Large Language Models (MLLMs).} Inspired by the powerful text generation capability of LLMs, researchers are exploring ways to extend the capabilities of LLMs to the visual domain, thus advancing the development of MLLMs \cite{alayrac2022flamingo, liu2023visual, chen2023minigpt, bai2023qwen}. Existing research on MLLMs aims to connect visual encoders with large language models. The visual encoder is employed to perceive visual information, and the LLM is used to understand semantic information and generate text. Flamingo \cite{alayrac2022flamingo} uses a visual encoder to extract visual embeddings, a resampler module to connect the visual encoder to a frozen language model, and multi-layer cross-attention module to fuse multi-modal features. BLIP2 \cite{li2023blip} uses the Q-Former module to connect the frozen LLM and visual encoder. MiniGPT-4 \cite{chen2023minigpt} and LLaVA \cite{liu2023visual} freeze the parameters of the vision encoder and LLM, and only optimize a trainable linear projector to connect the vision and language layers during the instruction tuning phase. 

\textbf{MLLMs for referring and grounding.} 
In recent investigations, works such as  Ferret \cite{you2023ferret}, GLaMM \cite{rasheed2023glamm} and Ferret-v2 \cite{zhang2024ferret}, closely resemble ours as they explore to enable MLLMs for image referring and grounding.  Ferret designs a spatial-aware visual sampler that supports free-form shapes of the referential region. For grounding, the coordinates of the bounding box are represented in natural language numerical form in the responses, followed by the corresponding text output. GLaMM supports pixel-level grounding segmentation and expands its vocabulary with special tokens to represent the input bounding box. Subsequently, the input tokens are replaced with RoI features extracted from the CLIP global image encoder layers, thereby enabling the use of specific regions as input. 

\textbf{MLLMs for Medical VQA.} 
Developments in multimodal large-scale language models have significantly impacted the biomedical field \cite{zhou2023skingpt, guo2023proteinchat, zhou2023path, xu2024mlevlm}. LLaVA-Med \cite{li2023llava} organizes biomedical image-text pairs from PubMed Central and uses GPT-4 \cite{chen2023sharegpt4v} to autonomously generate instruction data for multi-modal biomedical tasks. By constructing a large set of biomedical image and text pairs, LLaVA is fine-tuned to align image-text tags to the biomedical domain. Based on the framework OpenFlamingo \cite{awadalla2023openflamingo} and a pre-trained medical dataset, Med-Flamingo \cite{moor2023med} brings Flamingo's in-context learning capability into the medical field. This adaptation allows users to tailor response formats using few-shot prompts, such as asking for an explanation of an answer. HuatuoGPT-Vision \cite{chen2024huatuogpt} leverages the large-scale dataset PubMedVision containing 1.3 million samples to train the foundational MLLM. Furthermore, MedTrinity-25M \cite{xie2024medtrinity} employs an automated pipeline to introduce local annotations for regions of interest (ROIs), which include bounding boxes and segmentation masks. In the dataset construction strategy, different pre-trained models are utilized to annotate ROIs for the images, ensuring that each sample contains triplets of \{image, ROI, description \}. 

%PubMedVision consists of medical image-text pairs refined from PubMed and was reformatted with the assistance of GPT-4V for data denoising, producing a high-quality multimodal medical dataset containing 1.3 million samples.
\vspace{-0.2cm}
\section{Our Model: MIMO}

\subsection{Preliminary}
\label{sec:prelimiary}
\vspace{-0.2cm}
Given an input medical image $I \in \mathbb{R}^{H \times W \times 3}$ and an input query $Q$, our first goal is to obtain a response $R$ that contains a text response and seamlessly integrates segmentation masks directly tied to corresponding concepts. Let $T = [ t_1,t_2,...,t_n]$ be the text response in the form of a sequence of word tokens, then the final output $R$ will be:
\begin{align}
    R&=\left\{t_{j} \mid j=1,2, \ldots, n\right\} \nonumber \\ 
    &\cup\left\{<c_{i},s_{i}> \mid s_i \in S, c_i \in T,i=1,2, \ldots, m\right\},
\end{align}
where $\left\{t_{j} \mid j=1,2, \ldots, n\right\}$ denotes the text response $T$, and $C=\left\{c_{1}, c_{2}, \ldots, c_{m}\right\} \subseteq T$ is the set of medical concepts mentioned in the text response, $S = \left \{ s_1, s_2,...,s_m \right \}$ is the segmentation mask associated with each medical concept and $m  \le n$.

Optionally, additional visual prompts can be introduced to specify certain parts or concepts, helping to clarify the issue alongside the text. This means that the input query $Q$ has two forms: the text query $Q_{t}$ and the additional visual prompt $Q_{v}$.

\begin{figure*}
  \centering
  \includegraphics[width=0.8\textwidth]{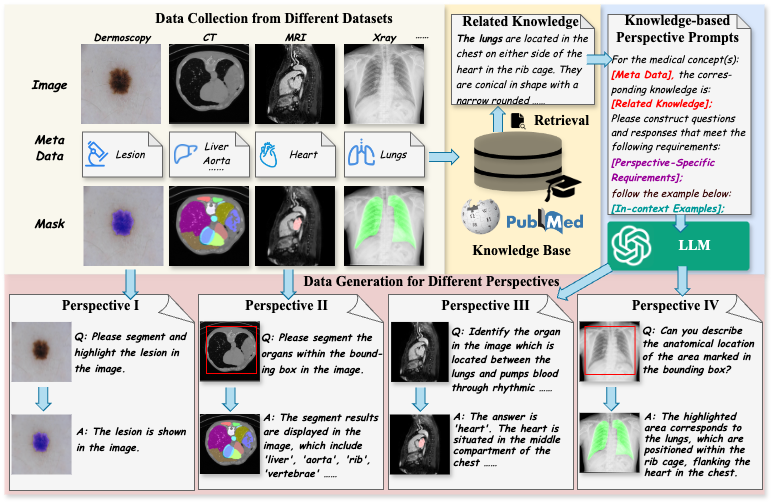} 
  \caption{The construction pipeline of MIMOSeg. The pipeline inclues data collection, knowledge retrieval, prompt construction and QA generation. The bottom of the figure shows example data from four perspectives of MIMOSeg.}
  \label{fig:dataset}
  \vspace{-0.5cm}
\end{figure*}

\subsection{Model Architecture}

As shown in Figure \ref{fig:model}, MIMO mainly consists of (i) an image encoder, (ii) a visual prompt encoder, (iii) a multi-modal input aligner, (iv) a LLM, (v) a segmentation vision encoder and (vi) a mask decoder. This model design accommodates both multimodal input and output formats, as defined in Section \ref{sec:prelimiary}.

\subsubsection{Visual Referring Multimodal Input}

For an input medical image $I \in \mathbb{R}^{H \times W \times 3}$, a CLIP visual encoder ViT-H/14 \cite{radford2021learning} is used, combined with a projection layer to produce the image embeddings $X_{img} \in \mathbb{R}^{l_1 \times d}$. For the input text query $Q_t$, the pre-trained tokenizer of LLM is used and projected to the text embeddings $X_q^t \in \mathbb{R}^{l_2 \times d}$. In addition, according to practical use, we consider two types of sparse visual prompts: points and bounding boxes. The prompt encoders adapt these additional visual prompts $Q_v$ to the same feature space as the image feature. Similar to the prompt encoder of SAM \cite{cheng2023sam}, points and boxes are encoded using positional encodings summed with learned embeddings for each prompt type. The corresponding embedding representation are then projected to $X_q^v \in \mathbb{R}^{l_3 \times d}$. The above image embeddings, text embeddings, and visual prompt embeddings will be concatenated into a unified multimodal feature.

\textbf{Multi-modal Input Aligner.} We design a multimodal input aligner to extract instruction-guided information from multi-modal features. The multi-modal input ($X_{img}$, $X_{q}^{t}$, $X_{q}^{v}$) interacts with the learnable query embedding $X_q$ through cross-attention, encouraging the extraction of image features that the textual and visual prompts focus on. Therefore, LLM receives visual information that helps follow textual and visual instructions. The output of Multi-modal input aligner contains the final learned instruction-guided query embedding $X_q$, which is then passed through a linear projection before being fed into the LLM.

% We demonstrate empirically (Table 2) that Multi-modal input aligner provides substantial performance improvements for both held-in and held-out evaluations. 
\vspace{-0.2cm}
\subsubsection{Language Grounding to Pixel Segmentation}

We use Vicuna \cite{zheng2023judging} as our language model, which is a decoder-only large language model based on LLaMA \cite{touvron2023llama}.
We introduce a pair of grounding tokens \texttt{<p>} and \texttt{<$/$p>} in the vocabulary of the model to indicate the start and end of groundable medical entities, and use a special token \texttt{<SEG>} immediately following the groundable entities to indicate potential segmentation requests. For example, a sentence in the final output $R$ from the model can be represented as ``\texttt{<p>}\textit{The central vein of the adrenal medulla}\texttt{<SEG>}\texttt{<$/$p>} \textit{is located in the }\texttt{<p>}\textit{adrenal medulla}\texttt{<SEG>}\texttt{<$/$p>} \textit{and is a rare type of blood vessel. Its structure is different from other veins, in which the }\texttt{<p>} \textit{smooth muscle} \texttt{<SEG>}\texttt{<$/$p>}\textit{ of the membrane is arranged in obvious longitudinal bundles.}''

In this way, the output embedding $r_{seg}$ corresponding to \texttt{<SEG>} tokens from the last layer of the LLM can be extracted and mapped to the feature space of the decoder through a projection layer. Then, the projected hidden states will be used by the mask decoder to generate binary segmentation masks $\mathcal{M}$. The process can be formulated as:
\begin{equation}
    \mathcal{M} = \mathcal{V}(\mathcal{G}(I), {proj}(r_{seg})),
\end{equation}
where $\mathcal{V}$ is the mask decoder, $\mathcal{G}$ is the segmentation mask encoder, and the input image will pass through $\mathcal{G}$ to obtain image features as additional input of the mask decoder and $proj$ represents the projection layer. We follow SAM to employ a pretrained SAM encoder as $\mathcal{G}$ and use the pretrained SAM decoder to instantiate $\mathcal{V}$.
\vspace{-0.2cm}
\subsection{Training Strategy}

We apply cross-entropy loss $\mathcal{L}_{text}$ for text generation. To compute the segmentation mask loss $\mathcal{L}_{mask}$, we use a combination of per-pixel binary cross-entropy (BCE) loss $\mathcal{L}_{bce}$ and DICE loss $\mathcal{L}_{dice}$. The final loss calculation is a linear combination of the language modeling loss $\mathcal{L}_{text}$ and these mask-related losses, which can be formulated as :
 \begin{equation}
     L_{total} = \lambda_1 L_{text} + \lambda_2 L_{bce} + \lambda_3 L_{dice},
 \end{equation}
where $\lambda_1,\lambda_2$ and $\lambda_3$ are the corresponding hyperparameters.

\section{Our Dataset: MIMOSeg}

\linespread{1.1}
\begin{table*}[htbp!]
    \caption{The results of comparison with previous studies on four perspectives in segmentations. Best and second-best results are shown in
\textbf{bold} and \underline{underline},  respectively. $\times$ means that the model does not support text instruction following.} 
    \fontsize{8}{8}\selectfont    %{字体尺寸}{行距}
    \tabcolsep=0.2cm %列间距
    \centering
    \begin{tabular}{c|rrrrrr}
    \toprule
    \multirow{1}{*}{Models}      
    & \multicolumn{1}{c}{\textbf{SAM-b} \cite{kirillov2023segment}}
    & \multicolumn{1}{c}{\textbf{SAM-h} \cite{kirillov2023segment}} 
    & \multicolumn{1}{c}{\textbf{SAM-Med} \cite{cheng2023sam}}
    & \multicolumn{1}{c|}{\textbf{GLaMM} \cite{rasheed2024glamm}}
    & \multicolumn{1}{c}{\textbf{MIMO}(w/o Aligner)}
    & \multicolumn{1}{c}{\textbf{MIMO}}\\
    \midrule
            
    \multicolumn{1}{c}{\textit{\textbf{Perspective I}}} 
    & \multicolumn{1}{c}{} 
    & \multicolumn{1}{c}{} 
    & \multicolumn{1}{c}{}
    & \multicolumn{1}{c}{} 
    & \multicolumn{1}{c}{}
    
    & \multicolumn{1}{c}{} \\     
    \midrule
        
    \multicolumn{1}{c|}{mIoU $\uparrow$} 
    & \multicolumn{1}{c}{$\times$} 
    & \multicolumn{1}{c}{$\times$} 
    & \multicolumn{1}{c}{$\times$}
    & \multicolumn{1}{c|}{0.556} 
    & \multicolumn{1}{c}{\underline{0.607}}
    & \multicolumn{1}{c}{\textbf{0.639}} \\

    \multicolumn{1}{c|}{AP50 $\uparrow$} 
    & \multicolumn{1}{c}{$\times$} 
    & \multicolumn{1}{c}{$\times$} 
    & \multicolumn{1}{c}{$\times$}
    & \multicolumn{1}{c|}{\textbf{0.324}} 
    & \multicolumn{1}{c}{0.301} 
    & \multicolumn{1}{c}{\underline{0.316}} \\

    \multicolumn{1}{c|}{F1 $\uparrow$} 
    & \multicolumn{1}{c}{$\times$} 
    & \multicolumn{1}{c}{$\times$} 
    & \multicolumn{1}{c}{$\times$}
    & \multicolumn{1}{c|}{\textbf{0.507}} 
    & \multicolumn{1}{c}{0.448} 
    & \multicolumn{1}{c}{\underline{0.457}} \\
    \midrule

    \multicolumn{1}{c}{\textit{\textbf{Perspective II}}} 
    & \multicolumn{1}{c}{} 
    & \multicolumn{1}{c}{} 
    & \multicolumn{1}{c}{}
    & \multicolumn{1}{c}{} 
    & \multicolumn{1}{c}{}
    
    & \multicolumn{1}{c}{} \\     
    \midrule
        
    \multicolumn{1}{c|}{mIoU $\uparrow$} 
    & \multicolumn{1}{c}{0.568} 
    & \multicolumn{1}{c}{0.571} 
    & \multicolumn{1}{c}{0.494}
    & \multicolumn{1}{c|}{0.496} 
    & \multicolumn{1}{c}{\underline{0.622}} 
    & \multicolumn{1}{c}{\textbf{0.665}} \\

    \multicolumn{1}{c|}{AP50 $\uparrow$} 
    & \multicolumn{1}{c}{0.125} 
    & \multicolumn{1}{c}{0.132} 
    & \multicolumn{1}{c}{0.097}
    & \multicolumn{1}{c|}{0.137} 
    & \multicolumn{1}{c}{\underline{0.185}} 
    & \multicolumn{1}{c}{\textbf{0.279}} \\

    \multicolumn{1}{c|}{F1 $\uparrow$} 
    & \multicolumn{1}{c}{$\times$} 
    & \multicolumn{1}{c}{$\times$} 
    & \multicolumn{1}{c}{$\times$}
    & \multicolumn{1}{c|}{0.279} 
    & \multicolumn{1}{c}{\underline{0.314}} 
    & \multicolumn{1}{c}{\textbf{0.406}} \\
    \midrule

    \multicolumn{1}{c}{\textit{\textbf{Perspective III}}} 
    & \multicolumn{1}{c}{} 
    & \multicolumn{1}{c}{} 
    & \multicolumn{1}{c}{}
    & \multicolumn{1}{c}{} 
    & \multicolumn{1}{c}{}
    
    & \multicolumn{1}{c}{} \\     
    \midrule
        
    \multicolumn{1}{c|}{mIoU $\uparrow$} 
    & \multicolumn{1}{c}{$\times$} 
    & \multicolumn{1}{c}{$\times$} 
    & \multicolumn{1}{c}{$\times$}
    & \multicolumn{1}{c|}{0.421} 
    & \multicolumn{1}{c}{\underline{0.468}} 
    & \multicolumn{1}{c}{\textbf{0.531}} \\

    \multicolumn{1}{c|}{AP50 $\uparrow$} 
    & \multicolumn{1}{c}{$\times$} 
    & \multicolumn{1}{c}{$\times$} 
    & \multicolumn{1}{c}{$\times$}
    & \multicolumn{1}{c|}{0.133} 
    & \multicolumn{1}{c}{\underline{0.205}} 
    & \multicolumn{1}{c}{\textbf{0.302}} \\

    \multicolumn{1}{c|}{F1 $\uparrow$} 
    & \multicolumn{1}{c}{$\times$} 
    & \multicolumn{1}{c}{$\times$} 
    & \multicolumn{1}{c}{$\times$}
    & \multicolumn{1}{c|}{0.253} 
    & \multicolumn{1}{c}{\underline{0.340}} 
    & \multicolumn{1}{c}{\textbf{0.427}} \\
    \midrule

    \multicolumn{1}{c}{\textit{\textbf{Perspective IV}}} 
    & \multicolumn{1}{c}{} 
    & \multicolumn{1}{c}{} 
    & \multicolumn{1}{c}{}
    & \multicolumn{1}{c}{} 
    & \multicolumn{1}{c}{}
    
    & \multicolumn{1}{c}{} \\     
    \midrule
        
    \multicolumn{1}{c|}{mIoU $\uparrow$} 
    & \multicolumn{1}{c}{0.573} 
    & \multicolumn{1}{c}{\underline{0.583}} 
    & \multicolumn{1}{c}{0.503}
    & \multicolumn{1}{c|}{0.564} 
    & \multicolumn{1}{c}{0.526} 
    & \multicolumn{1}{c}{\textbf{0.586}} \\

    \multicolumn{1}{c|}{AP50 $\uparrow$} 
    & \multicolumn{1}{c}{\underline{0.363}}
    & \multicolumn{1}{c}{\textbf{0.404}} 
    & \multicolumn{1}{c}{0.300}
    & \multicolumn{1}{c|}{0.267} 
    & \multicolumn{1}{c}{0.253} 
    & \multicolumn{1}{c}{0.309} \\

    \multicolumn{1}{c|}{F1 $\uparrow$} 
    & \multicolumn{1}{c}{$\times$} 
    & \multicolumn{1}{c}{$\times$} 
    & \multicolumn{1}{c}{$\times$}
    & \multicolumn{1}{c|}{\underline{0.422}} 
    & \multicolumn{1}{c}{0.401} 
    & \multicolumn{1}{c}{\textbf{0.470}} \\
    
    \bottomrule
    \end{tabular}
    \label{tab:held-in-seg}
    \vspace{-0.3cm}
\end{table*}

%To support the capability of MIMO in visual referring and pixel grounding, we propose the MIMOSeg dataset, a diverse and comprehensive collection of annotated medical images with segmentation masks. Designed to span multiple imaging modalities and different types of prompts, MIMOSeg serves as a foundational resource. By offering high-quality annotations and diverse clinical scenarios, MIMOSeg enables the development of robust medical vision-language models capable of precise and interpretable segmentation.

% \subsection{Datasource}
% \label{datasource}
% Datasets for pixel-level grounding tasks need to contain a large number of medical images and segmentation labels from different modalities. To address these difficulties, our proposed MIMOSeg Dataset is constructed based on a wide range of publicly available segmentation datasets, with about 1 million medical pixel-level grounding samples, including data from 8 different modalities such as CT, X-rays, fundus images, and pathology sections. The data source distribution is shown in Figure \ref{}. More details can be found in Appendix \ref{}.

\subsection{Construction Perspectives}

To guide the construction of the MIMOSeg dataset, we summarize four key perspectives, each aiming to enhance the utility of the dataset in various pixel-level grounding tasks.

\textbf{Perspective I: Language-Guided Segmentation. }This perspective represents the basic instruction-following ability of the model. The model receives an image and a language instruction, producing segmentation masks, with no visual prompt provided. From this perspective, the model needs to build the ability to recognize medical entities and achieve the mapping from text to segmentation. An example instruction is: ``\textit{Please segment out the heart in the picture.}''

\textbf{Perspective II: Visual Prompt Perceiving.} This perspective represents the basic visual prompt perception ability of the model. The model receives an image, a language instruction, and a visual prompt (box or point), producing segmentation masks. From this perspective, the model needs to perceive and understand the given visual prompts, comprehend the semantics of visual entities, and establish a mapping from segmentation to text. An example instruction is: ``\textit{Please segment all the organs in the box in the picture.}''

\textbf{Perspective III: Responses with Segmentation Aligning.} This perspective deals with complex tasks involving question-answering. The model is required to reason about the image and generate a mask that aligns with both the question and visual segmentation, blending language understanding with pixel-level image analysis. From this perspective, the capabilities of the model are more challenging. An example question is: ``\textit{Could you identify the organ in the image that has rhythmic sequences of contractions that propagate through the atrioventricular node and conduction system?}''

\textbf{Perspective IV: Visual Prompt Assisted Questioning. } This perspective deals with answering a question with the assistance of a visual prompt. The visual prompt aids in directing attention, while the model needs to generate both an answer and a corresponding segmentation mask based on the question. From this perspective, the model needs to incorporate visual prompts to understand the question based on complex reasoning. An example question is: ``\textit{Can you provide details about the physiological functions of the area within the bounding box that appears to be conical and situated in the chest?}''

The four perspectives in the construction of the MIMOSeg Dataset provide a comprehensive framework for addressing different aspects of pixel-level grounding tasks. The ground-segmentation perspectives (I \& II) allow models to focus on accurately delineating anatomical structures based on either text or visual cues, while the QA-driven perspectives (III \& IV) simulate more complex, interactive clinical scenarios, where both segmentation and reasoning are required. This multi-perspective approach enriches the applicability of the dataset and guides the subsequent construction pipeline.

\subsection{Construction Pipeline}
The complete construction pipeline is illustrated in Figure \ref{fig:dataset}, which includes data collection, knowledge retrieval, prompt construction and Q\&A generation.

\textbf{Data Source. }In order to solve the problems under the above perspectives, our proposed MIMOSeg Dataset is constructed based on a wide range of publicly available segmentation datasets, with about 1 million medical pixel-level grounding samples, including data from 8 different modalities such as CT, X-rays, fundus images, and pathology sections. More details can be found in Appendix {\color{red} B.1}. We divide these datasets to meet the construction needs of different perspectives. For each perspective, we select a sufficient number of single-label and multi-label datasets, ensuring comprehensive coverage of image modalities. The specific division is detailed in Appendix {\color{red} B.3}.

\textbf{Knowledge Retrieval.} Unlike general fields, the medical field is highly dependent on knowledge, and most medical entities have corresponding definitions. Therefore, we manually constructed a knowledge base that covers all segmentation labels via some easy-access knowledge sources $\footnote{Wikipedia \url{https://en.wikipedia.org/wiki/}}$ \footnote{UMLS \url{https://www.nlm.nih.gov/research/umls/}}. The knowledge includes common information on disease causes, treatment methods, common symptoms, as well as the locations and functions of organs. For the metadata of the input data, we can retrieve corresponding knowledge for each label from the knowledge base.

\textbf{Data Generation for different perspectives.} We directly construct corresponding instruction templates and response templates for Perspective I \& II. For Perspective III \& IV, we design knowledge-based perspective prompts to make GPT-4o \cite{achiam2023gpt} generate meaningful questions and answers in a tone as if it could see the image (even though
it only has access to the text). In the prompt, each perspective will correspond to different perspective-specific requirements. Perspective III focuses on identifying organs or lesions through complex questions and providing analytical reasons, while Perspective IV emphasizes visual analysis and content understanding based on visual prompts. The knowledge corresponding to the label of each image in the aforementioned knowledge base will be input into the prompt.  For multi-label tasks, the generated questions and answers must consider each label and their interrelationships. We also manually construct relevant in-context examples to enhance the generation of Q\&A pairs. The specific prompts are detailed in Appendix {\color{red} E.2}. Examples of Q\&A content for the four types of perspectives are shown in Appendix {\color{red} C}.

% \textbf{Prompt Construction. } For Perspective III and Perspective IV, we use GPT-4o\cite{achiam2023gpt} for data construction. For the retrieved knowledge of several labels, we splice the retrieved relevant knowledge into the prompt. In the prompt, each perspective will correspond to different perspective-specific requirements. Perspective III focuses on identifying organs or lesions through complex questions and providing analytical reasons, while Perspective IV emphasizes visual analysis and content understanding based on visual prompts. For multi-label tasks, the generated questions and answers must consider each label and their interrelationships. We also manually constructed relevant in-context examples to enhance the generation of Q\&A pairs. The specific prompts are detailed in Appendix \ref{appendix:prompts_III_IV}.

% \textbf{QA Generation.} For Perspective I and Perspective II, we directly construct corresponding instruction templates and response templates. The content of the templates can be found in \ref{appendix:instructions_I_II}. For Perspective III and Perspective IV, we use the Q\&A pairs generated by GPT-4o and review the formatting. Examples of QA content for the four types of perspectives are shown in Appendix \ref{appendix:data_analysis}.
\section{Experiments and Analysis}

\linespread{1.1}
\begin{table*}[htbp!]
    \caption{The results of comparison with previous studies on four perspectives in question answering. Best and second-best results are shown in \textbf{bold} and \underline{underline},  respectively. } 
    \fontsize{8}{8}\selectfont    %{字体尺寸}{行距}
    \tabcolsep=0.2cm %列间距
    \centering
    \begin{tabular}{c|rrrrrr}
    \toprule
    \multirow{1}{*}{Models}      
    & \multicolumn{1}{c}{\textbf{GPT-4o} \cite{achiam2023gpt}}
    & \multicolumn{1}{c}{\textbf{LLaVA-Med} \cite{li2023llava}} 
    & \multicolumn{1}{c}{\textbf{HuatuoGPT-Vision} \cite{chen2024huatuogpt}}
    & \multicolumn{1}{c|}{\textbf{GLaMM} \cite{rasheed2024glamm}}
    & \multicolumn{1}{c}{\textbf{MIMO}(w/o Aligner)}
    & \multicolumn{1}{c}{\textbf{MIMO}}\\
    \midrule
            
    \multicolumn{1}{c}{\textit{\textbf{Perspective II}}} 
    & \multicolumn{1}{c}{} 
    & \multicolumn{1}{c}{} 
    & \multicolumn{1}{c}{}
    & \multicolumn{1}{c}{} 
    & \multicolumn{1}{c}{}
    
    & \multicolumn{1}{c}{} \\     
    \midrule
        
    \multicolumn{1}{c|}{ROUGE-L $\uparrow$} 
    & \multicolumn{1}{c}{0.169} 
    & \multicolumn{1}{c}{0.120} 
    & \multicolumn{1}{c}{0.070}
    & \multicolumn{1}{c|}{\textbf{0.436}} 
    & \multicolumn{1}{c}{0.317} 
    & \multicolumn{1}{c}{\underline{0.335}} \\

    \multicolumn{1}{c|}{BLEU-4 $\uparrow$} 
    & \multicolumn{1}{c}{0.002} 
    & \multicolumn{1}{c}{0.005} 
    & \multicolumn{1}{c}{0.001}
    & \multicolumn{1}{c|}{\textbf{0.293}} 
    & \multicolumn{1}{c}{0.171} 
    & \multicolumn{1}{c}{\underline{0.181}} \\

    \multicolumn{1}{c|}{METEOR $\uparrow$} 
    & \multicolumn{1}{c}{0.053} 
    & \multicolumn{1}{c}{0.077} 
    & \multicolumn{1}{c}{0.062}
    & \multicolumn{1}{c|}{\textbf{0.304}} 
    & \multicolumn{1}{c}{0.208} 
    & \multicolumn{1}{c}{\underline{0.253}} \\

    % \multicolumn{1}{c|}{Bert-Sim $\uparrow$} 
    % & \multicolumn{1}{c}{$\times$} 
    % & \multicolumn{1}{c}{$\times$} 
    % & \multicolumn{1}{c}{$\times$} 
    % & \multicolumn{1}{c|}{0.573} 
    % & \multicolumn{1}{c}{0.363} 
    % & \multicolumn{1}{c}{$\times$} \\
    \midrule

    \multicolumn{1}{c}{\textit{\textbf{Perspective III}}} 
    & \multicolumn{1}{c}{} 
    & \multicolumn{1}{c}{} 
    & \multicolumn{1}{c}{}
    & \multicolumn{1}{c}{} 
    & \multicolumn{1}{c}{}
    
    & \multicolumn{1}{c}{} \\     
    \midrule
        
    \multicolumn{1}{c|}{ROUGE-L $\uparrow$} 
    & \multicolumn{1}{c}{0.146} 
    & \multicolumn{1}{c}{0.269} 
    & \multicolumn{1}{c}{0.278}
    & \multicolumn{1}{c|}{0.339} 
    & \multicolumn{1}{c}{\textbf{0.650}} 
    & \multicolumn{1}{c}{\underline{0.581}} \\

    \multicolumn{1}{c|}{BLEU-4 $\uparrow$} 
    & \multicolumn{1}{c}{0.007} 
    & \multicolumn{1}{c}{0.129} 
    & \multicolumn{1}{c}{0.064}
    & \multicolumn{1}{c|}{0.150} 
    & \multicolumn{1}{c}{\textbf{0.564}} 
    & \multicolumn{1}{c}{\underline{0.475}} \\

    \multicolumn{1}{c|}{METEOR $\uparrow$} 
    & \multicolumn{1}{c}{0.073} 
    & \multicolumn{1}{c}{0.163} 
    & \multicolumn{1}{c}{0.211}
    & \multicolumn{1}{c|}{0.198} 
    & \multicolumn{1}{c}{\textbf{0.392}} 
    & \multicolumn{1}{c}{\underline{0.334}} \\

    % \multicolumn{1}{c|}{Bert-Sim $\uparrow$} 
    % & \multicolumn{1}{c}{$\times$} 
    % & \multicolumn{1}{c}{$\times$} 
    % & \multicolumn{1}{c}{$\times$} 
    % & \multicolumn{1}{c|}{0.573} 
    % & \multicolumn{1}{c}{0.363} 
    % & \multicolumn{1}{c}{$\times$} \\
    \midrule

    \multicolumn{1}{c}{\textit{\textbf{Perspective IV}}} 
    & \multicolumn{1}{c}{} 
    & \multicolumn{1}{c}{} 
    & \multicolumn{1}{c}{}
    & \multicolumn{1}{c}{} 
    & \multicolumn{1}{c}{}
    
    & \multicolumn{1}{c}{} \\     
    \midrule
        
    \multicolumn{1}{c|}{ROUGE-L $\uparrow$} 
    & \multicolumn{1}{c}{0.223} 
    & \multicolumn{1}{c}{0.215} 
    & \multicolumn{1}{c}{0.222}
    & \multicolumn{1}{c|}{0.380} 
    & \multicolumn{1}{c}{\underline{0.387}} 
    & \multicolumn{1}{c}{\textbf{0.397}} \\

    \multicolumn{1}{c|}{BLEU-4 $\uparrow$} 
    & \multicolumn{1}{c}{0.061} 
    & \multicolumn{1}{c}{0.046} 
    & \multicolumn{1}{c}{0.034}
    & \multicolumn{1}{c|}{0.240} 
    & \multicolumn{1}{c}{\textbf{0.263}} 
    & \multicolumn{1}{c}{\textbf{0.263}} \\

    \multicolumn{1}{c|}{METEOR $\uparrow$} 
    & \multicolumn{1}{c}{0.159} 
    & \multicolumn{1}{c}{0.127} 
    & \multicolumn{1}{c}{0.156}
    & \multicolumn{1}{c|}{0.216} 
    & \multicolumn{1}{c}{\underline{0.218}} 
    & \multicolumn{1}{c}{\textbf{0.222}} \\

    % \multicolumn{1}{c|}{Bert-Sim $\uparrow$} 
    % & \multicolumn{1}{c}{$\times$} 
    % & \multicolumn{1}{c}{$\times$} 
    % & \multicolumn{1}{c}{$\times$} 
    % & \multicolumn{1}{c|}{0.573} 
    % & \multicolumn{1}{c}{0.363} 
    % & \multicolumn{1}{c}{$\times$} \\
    
    \bottomrule
    \end{tabular}
    \label{tab:held-in-cap}
    \vspace{-0.3cm}
\end{table*}

\linespread{1.1}
\begin{table*}[htbp!]
    \caption{Comparison with previous work on the held-out segmentation datasets. Best and second-best results are shown in
\textbf{bold} and \underline{underline},  respectively. $\times$ means that the model does not support text instruction following.} 
    \fontsize{8}{10}\selectfont    %{字体尺寸}{行距}
    \centering
    \begin{tabular}{c|rrrr|rrrr|rrrr}
    \toprule
    \multirow{4}{*}{Models}      
    & \multicolumn{4}{c|}{\textbf{X-ray}}
    & \multicolumn{4}{c|}{\textbf{Fundus}} 
    & \multicolumn{4}{c}{\textbf{Skin Lesion}}\\
    \cmidrule{2-13} 

    & \multicolumn{2}{c|}{w/o bbox} 
    & \multicolumn{2}{c|}{with bbox} 
    & \multicolumn{2}{c|}{w/o bbox} 
    & \multicolumn{2}{c|}{with bbox} 
    & \multicolumn{2}{c|}{w/o bbox} 
    & \multicolumn{2}{c}{with bbox} \\
    \cmidrule{2-13}
 
    & \multicolumn{1}{c}{AP50} 
    & \multicolumn{1}{c|}{mIoU}
    & \multicolumn{1}{c}{AP50} 
    & \multicolumn{1}{c|}{mIoU}
    & \multicolumn{1}{c}{AP50} 
    & \multicolumn{1}{c|}{mIoU}
    & \multicolumn{1}{c}{AP50} 
    & \multicolumn{1}{c|}{mIoU}
    & \multicolumn{1}{c}{AP50} 
    & \multicolumn{1}{c|}{mIoU}
    & \multicolumn{1}{c}{AP50} 
    & \multicolumn{1}{c}{mIoU}\\
    \midrule
            
    % \multicolumn{1}{c}{\textit{\textbf{Zero-shot with SAMs}}} 
    % & \multicolumn{1}{c}{} 
    % & \multicolumn{1}{c}{} 
    % & \multicolumn{1}{c}{}
    % & \multicolumn{1}{c}{} 
    % & \multicolumn{1}{c}{}
    
    % & \multicolumn{1}{c}{} 
    % & \multicolumn{1}{c}{} 
    % & \multicolumn{1}{c}{}
    % & \multicolumn{1}{c}{} \\     
    % \midrule
        
    \multicolumn{1}{c|}{SAM-b \cite{kirillov2023segment}} 
    & \multicolumn{1}{c}{$\times$} 
    & \multicolumn{1}{c|}{$\times$}
    & \multicolumn{1}{c}{0.971} 
    & \multicolumn{1}{c|}{0.855}
    & \multicolumn{1}{c}{$\times$} 
    & \multicolumn{1}{c|}{$\times$}
    & \multicolumn{1}{c}{0.923} 
    & \multicolumn{1}{c|}{0.766}
    & \multicolumn{1}{c}{$\times$} 
    & \multicolumn{1}{c|}{$\times$}
    & \multicolumn{1}{c}{0.952} 
    & \multicolumn{1}{c}{0.851}\\
           
    \multicolumn{1}{c|}{SAM-h \cite{kirillov2023segment}} 
    & \multicolumn{1}{c}{$\times$} 
    & \multicolumn{1}{c|}{$\times$}
    & \multicolumn{1}{c}{\textbf{0.989}} 
    & \multicolumn{1}{c|}{0.835}
    & \multicolumn{1}{c}{$\times$} 
    & \multicolumn{1}{c|}{$\times$}
    & \multicolumn{1}{c}{\underline{0.973}} 
    & \multicolumn{1}{c|}{0.865}
    & \multicolumn{1}{c}{$\times$} 
    & \multicolumn{1}{c|}{$\times$}
    & \multicolumn{1}{c}{{0.982}} 
    & \multicolumn{1}{c}{\underline{0.861}}\\

    \multicolumn{1}{c|}{SAM-Med \cite{cheng2023sam}} 
    & \multicolumn{1}{c}{$\times$} 
    & \multicolumn{1}{c|}{$\times$}
    & \multicolumn{1}{c}{0.955} 
    & \multicolumn{1}{c|}{0.877}
    & \multicolumn{1}{c}{$\times$} 
    & \multicolumn{1}{c|}{$\times$}
    & \multicolumn{1}{c}{0.678} 
    & \multicolumn{1}{c|}{0.662}
    & \multicolumn{1}{c}{$\times$} 
    & \multicolumn{1}{c|}{$\times$}
    & \multicolumn{1}{c}{0.931} 
    & \multicolumn{1}{c}{0.848}\\
    % \midrule
            
    % \multicolumn{1}{c}{\textit{\textbf{Fine-tuning with general VLMs}}} 
    % & \multicolumn{1}{c}{} 
    % & \multicolumn{1}{c}{} 
    % & \multicolumn{1}{c}{}
    % & \multicolumn{1}{c}{} 
    % & \multicolumn{1}{c}{}
    
    % & \multicolumn{1}{c}{} 
    % & \multicolumn{1}{c}{} 
    % & \multicolumn{1}{c}{}
    % & \multicolumn{1}{c}{} \\     
    % \midrule
            
    % \multicolumn{1}{c|}{NextChat \cite{li2023llavamed}} 
    % & \multicolumn{1}{c}{AP50} 
    % & \multicolumn{1}{c|}{mIoU}
    % & \multicolumn{1}{c}{AP50} 
    % & \multicolumn{1}{c|}{mIoU}
    % & \multicolumn{1}{c}{AP50} 
    % & \multicolumn{1}{c|}{mIoU}
    % & \multicolumn{1}{c}{AP50} 
    % & \multicolumn{1}{c|}{mIoU}
    % & \multicolumn{1}{c}{AP50} 
    % & \multicolumn{1}{c|}{mIoU}
    % & \multicolumn{1}{c}{AP50} 
    % & \multicolumn{1}{c}{mIoU}\\
           
    \multicolumn{1}{c|}{GLaMM \cite{rasheed2024glamm}} 
    & \multicolumn{1}{c}{0.335} 
    & \multicolumn{1}{c|}{0.554}
    & \multicolumn{1}{c}{0.859} 
    & \multicolumn{1}{c|}{\underline{0.882}}
    & \multicolumn{1}{c}{\textbf{0.940}} 
    & \multicolumn{1}{c|}{\textbf{0.870}}
    & \multicolumn{1}{c}{0.946} 
    & \multicolumn{1}{c|}{0.874}
    & \multicolumn{1}{c}{{0.458}} 
    & \multicolumn{1}{c|}{{0.526}}
    & \multicolumn{1}{c}{0.719} 
    & \multicolumn{1}{c}{0.723}\\

    \midrule
          
    % \multicolumn{1}{c}{\textit{\textbf{Our methods}}} 
    % & \multicolumn{1}{c}{} 
    % & \multicolumn{1}{c}{} 
    % & \multicolumn{1}{c}{}
    % & \multicolumn{1}{c}{}
    % & \multicolumn{1}{c}{}
    
    % &\multicolumn{1}{c}{} 
    % &\multicolumn{1}{c}{}
    % & \multicolumn{1}{c}{}
    % & \multicolumn{1}{c}{}\\
    % \midrule

    \rowcolor{gray!10} 
    \multicolumn{1}{c|}{MIMO (w/o Aligner)} 
    & \multicolumn{1}{c}{0.313} 
    & \multicolumn{1}{c|}{0.490}
    & \multicolumn{1}{c}{0.822} 
    & \multicolumn{1}{c|}{0.813}
    & \multicolumn{1}{c}{0.915} 
    & \multicolumn{1}{c|}{0.852}
    & \multicolumn{1}{c}{0.951} 
    & \multicolumn{1}{c|}{\textbf{0.883}}
    & \multicolumn{1}{c}{\underline{0.747}} 
    & \multicolumn{1}{c|}{\underline{0.638}}
    & \multicolumn{1}{c}{\textbf{0.985}} 
    & \multicolumn{1}{c}{\underline{0.861}}\\
          
    \rowcolor{gray!10}     
    \multicolumn{1}{c|}{MIMO} 
    & \multicolumn{1}{c}{\textbf{0.507}} 
    & \multicolumn{1}{c|}{\textbf{0.718}}
    & \multicolumn{1}{c}{\textbf{0.989}} 
    & \multicolumn{1}{c|}{\textbf{0.883}}
    & \multicolumn{1}{c}{\textbf{0.940}} 
    & \multicolumn{1}{c|}{\underline{0.863}}
    & \multicolumn{1}{c}{\textbf{0.988}} 
    & \multicolumn{1}{c|}{\underline{0.881}}
    & \multicolumn{1}{c}{\textbf{0.787}} 
    & \multicolumn{1}{c|}{\textbf{0.647}}
    & \multicolumn{1}{c}{\textbf{0.985}} 
    & \multicolumn{1}{c}{\textbf{0.871}}\\
    \bottomrule
    \end{tabular}
    \label{tab:held-out-seg}
    \vspace{-0.5cm}
\end{table*}

\subsection{Experimental Setup}

\textbf{Datasets.} Since MIMOSeg is the first dataset with both visual referring and pixel grounding, we conduct held-in experiments on the MIMOSeg test set. We show experimental results under four perspectives to compare the capabilities of the models more reasonably. Meanwhile, to evaluate the generalization ability of our model, we select 6 held-out datasets for zero-shot experiments, including 3 medical VQA datasets Slake \cite{liu2021slake}, RadVQA \cite{lau2018dataset}, PathVQA \cite{he2020pathvqa} and 3 medical segmentation datasets X-ray \cite{cheng2023sam}, Fundus \cite{cheng2023sam}, Skin Lesion \cite{rajinikanth2019skin}. 

\textbf{Metrics.} In the held-in experiments, we report BLEU-4 \cite{papineni2002bleu}, ROUGE-L \cite{rouge2004package}, METEOR \cite{banerjee2005meteor}  for text generation evaluation, AP50 and mIOU for mask segmentation, and F1 Score for mask-to-entity correspondence accuracy (refer to Appendix {\color{red} D} for details). In the held-out experiments, for the VQA task, we evaluate the accuracy of the model's responses. The experimental settings remain the same as HuatuoGPT-Vision. For the segmentation task, we use the same settings as the held-in experiments.
% for the evaluation of generated text, we use traditional text generation evaluation metrics, including BLEU-4, ROUGE-L, METEOR; for the evaluation of mask segmentation, we report AP50 and mIOU. To measure the correspondence between the generated labels and the generated mask, we use F1 Score for evaluation.

\textbf{Implementation Details.} We use Vicuna LLM with 7B parameters as the default large language model, and the multi-modal input aligner is randomly initialized. After initialization, MIMO is trained on the aforementioned MIMOSeg and \text{LLaVA-Med} VQA \cite{li2023llava} dataset. Specifically, the four perspectives of MIMOSeg and the \text{LLaVA-Med} VQA dataset are mixed for training with a ratio of $1:2:2:1:1$ for $3$ epochs. More implementation details can be found in Appendix {\color{red} D}.

\subsection{Comparison with Previous Studies}

We conduct comparative experiments of our method with existing SAMs, existing medical VLMs, and general VLMs. For existing SAMs and medical VLMs, we directly conduct zero-shot experiments. For general VLMs, we select a base model that supports multimodal input and output, and fine-tune it using MIMOSeg. It should be noted that since there is no model in the general field combines both visual reference multimodal input and pixel-grounded multimodal output, we only select the most similar model and make appropriate modifications to better suit our task. The experimental results are shown in Table \ref{tab:held-in-seg} to Table \ref{tab:held-out-cap}. 

\textbf{Held-in experiments.} 
We first evaluate MIMO on the held-in dataset, summarizing results from four different perspectives in Table \ref{tab:held-in-seg} and Table \ref{tab:held-in-cap}.  For the segmentation performance in tasks involving visual prompts (Perspective II \& IV), we compare MIMO’s performance with SAM and SAM-Med. Since SAM only supports segmentation mask outputs and does not provide textual responses, its F1 scores are marked as $\times$. As shown in Table \ref{tab:held-in-seg}, MIMO demonstrates significantly better performance compared to baseline methods.

\begin{figure*}[htbp!]
  \centering
  \includegraphics[width=1\textwidth]{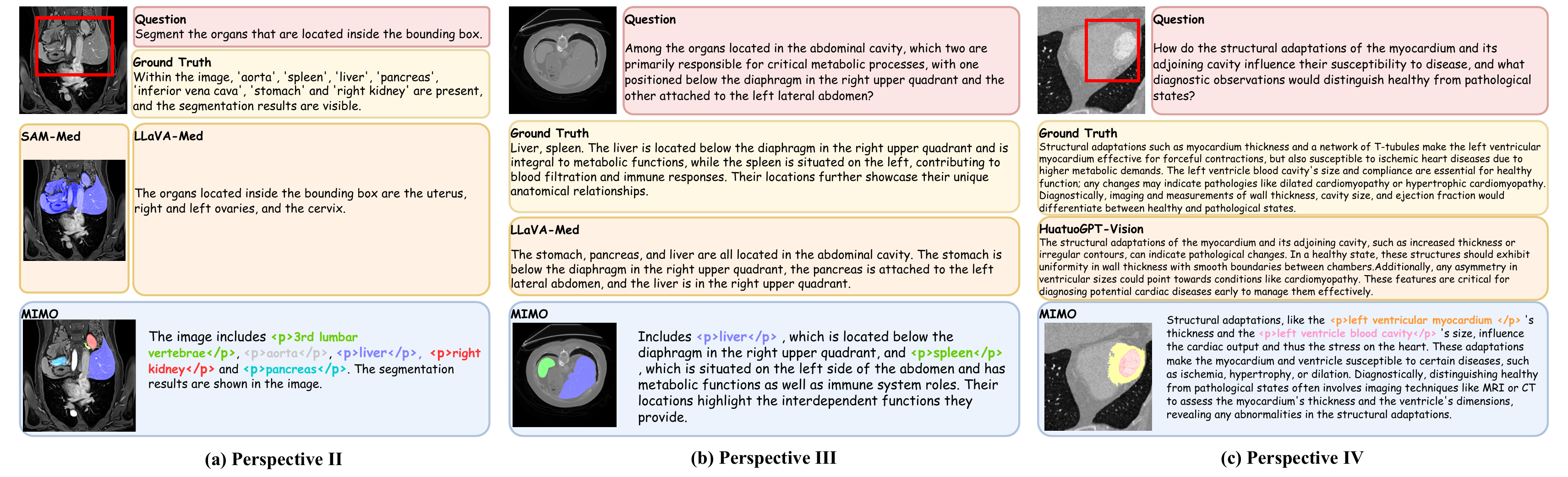} 
  \vspace{-0.9cm}
  \caption{Qualitative analysis of experimental results from three perspectives. We compare the results with LLaVA-Med and HuatuoGPT-Vision. MIMO can generate segmentation masks with relevant medical entities while outputting the answer.}
  \label{fig:case}
  \vspace{-0.5cm}
\end{figure*}

In complex scenarios (Perspective III \& IV), MIMO outperforms our reproduced GLaMM model, indicating the architectural advantages of MIMO.  We also present experimental results using medical VLMs and GPT-4o to evaluate the text responses. Given that MIMOSeg's Q\&A design focuses on specific entities within images, the four tasks based on distinct perspectives are challenging for traditional VLMs (refer to Figure \ref{fig:case} and Appendix {\color{red} F} for qualitative analysis). The enhanced evaluation results of MIMO further highlight the significance of integrating unified referring and grounding capabilities in MVQA models.

\textbf{Held-out experiments.} Table \ref{tab:held-out-seg} and Table \ref{tab:held-out-cap} show the held-out experimental results. In the segmentation task, we perform segmentation with and without bounding boxes. Since SAM and SAM-Med only support visual prompt input, their results in the table w/o bbox is $\times$. From Table \ref{tab:held-out-seg}, we can see that MIMO achieves the best results in both types of segmentation. The experimental results show that MIMO has good generalization ability in the case of text instruction segmentation; in the case of segmentation with visual prompts, MIMO's segmentation effect also outperforms the previous SAM and SAM-Med models.
In the VQA task, we conduct experiments on 3 public benchmarks. We choose closed-ended questions in the benchmarks to more accurately evaluate the accuracy of the generated results. Among them, HuatuoGPT-Vision uses more datasets for training. Compared with the remaining models, our model achieves the best accuracy on Slake and VQA-RAD. At the same time, MIMO also has the highest average accuracy on the three datasets.

% The experimental settings remain the same as HuatuoGPT-Vision.

\subsection{Ablation Studies}

\textbf{Ablation of Multi-modal Input Aligner.} 
Comparing the results of MIMO and MIMO (w/o aligner) in Table \ref{tab:held-in-seg} to Table \ref{tab:held-out-seg}, MIMO shows higher results in most cases, especially in the case of inputs with bounding boxes such as Perspective II \& IV. These results prove the effectiveness of our designed aligner. The multimodal aligner can get more accurate content understanding when the input has visual prompts.

\begin{figure}
  \centering
  \includegraphics[width=1\linewidth]{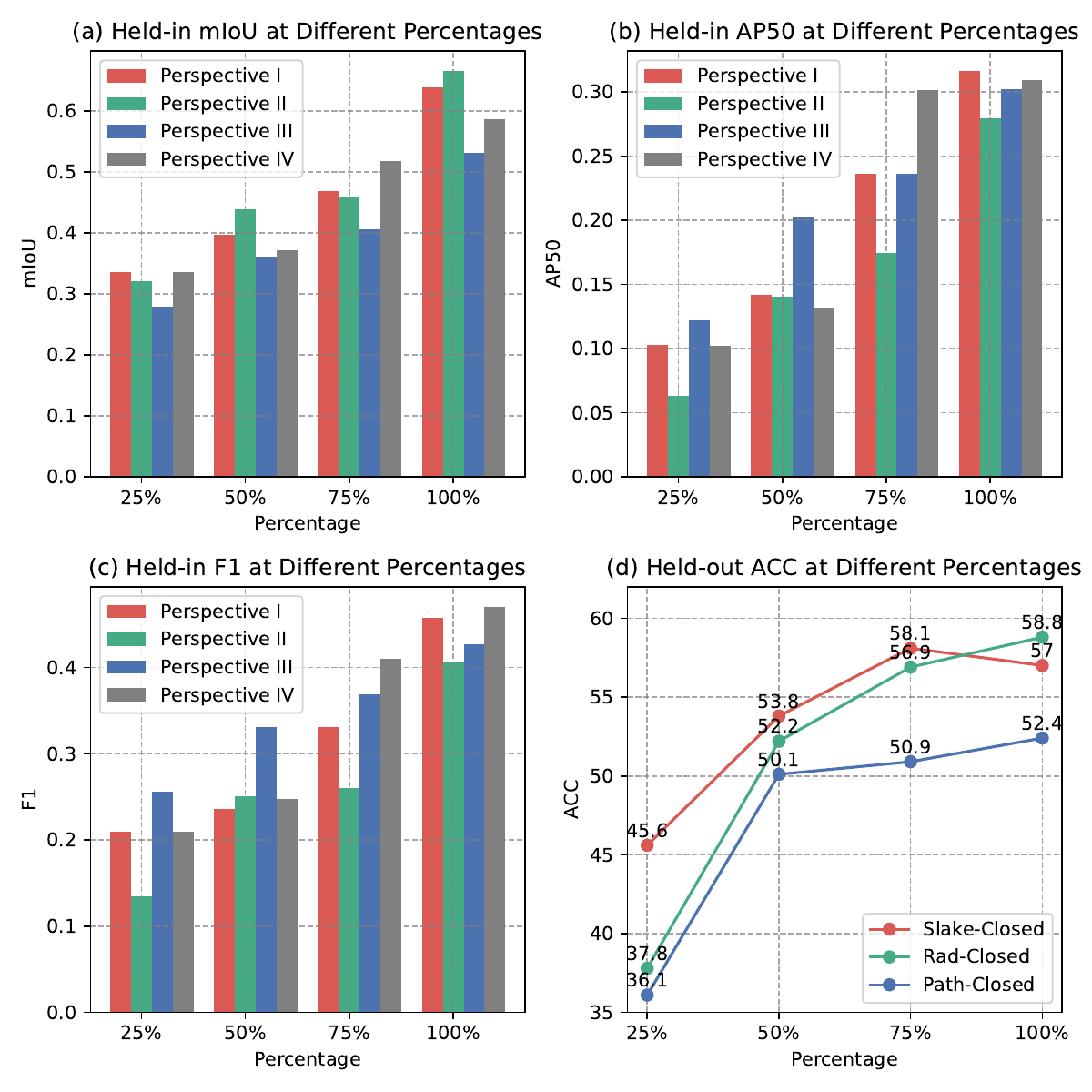} 
  \vspace{-0.8cm}
  \caption{Held-in and held-out experimental results with different ratios of training data.}
  \label{fig:abliation}
  \vspace{-0.6cm}
\end{figure}

\textbf{Ablation of Training Data Ratio.} 
The amount of data is an important factor affecting the experiment. To examine the impact of data quantity on model performance, we compare the evaluation results under different data ratios. Experiments are conducted using 25\%, 50\%, 75\% and 100\% of the training data. The experimental results are shown in Figure \ref{fig:abliation}. Observing Figures \ref{fig:abliation}(a), (b) and (c), we can find that as the amount of data increases, the model's mIoU, AP50, F1 and other indicators are steadily increasing. In Figure \ref{fig:abliation}(d), for VQA-RAD and PathVQA, the result also shows a similar trend. These experimental results prove the positive correlation between the model effect and the amount of data, and also prove that the MIMOSeg dataset can bring positive effects to multimodal medical tasks.

\linespread{1.1}
\begin{table}[htbp!]
    \caption{Comparison with previous work on the held-out medical visual question answering datasets.} 
    \fontsize{8}{10}\selectfont    %{字体尺寸}{行距}
    \tabcolsep=0.1cm %列间距
    \centering
    \begin{tabular}{c|rrr|r}
    \toprule
    \multicolumn{1}{c|}{Models}      
    & \multicolumn{1}{c}{\textbf{VQA-RAD}} 
    & \multicolumn{1}{c}{\textbf{SLAKE}}
    & \multicolumn{1}{c|}{\textbf{PathVQA}}
    & \multicolumn{1}{c}{\textbf{Avg.}}\\
    
    \midrule
    \multicolumn{1}{c|}{\color{gray}HuatuoGPT-Vision \cite{chen2024huatuogpt}} 
    & \multicolumn{1}{c}{\color{gray}\textbf{63.8}} 
    & \multicolumn{1}{c}{\color{gray}\textbf{74.5}}
    & \multicolumn{1}{c|}{\color{gray}\textbf{59.9}}
    & \multicolumn{1}{c}{\color{gray}66.1}\\ 

    \multicolumn{1}{c|}{Med-Flamingo \cite{moor2023med}} 
    & \multicolumn{1}{c}{45.4} 
    & \multicolumn{1}{c}{43.5}
    & \multicolumn{1}{c|}{54.7}
    & \multicolumn{1}{c}{47.9}\\ 

    \multicolumn{1}{c|}{RAD-FM \cite{wu2023towards}} 
    & \multicolumn{1}{c}{50.6} 
    & \multicolumn{1}{c}{34.6}
    & \multicolumn{1}{c|}{38.7}
    & \multicolumn{1}{c}{41.3}\\ 
           
    \multicolumn{1}{c|}{LLaVa-Med \cite{li2023llava}} 
    & \multicolumn{1}{c}{51.4} 
    & \multicolumn{1}{c}{48.6}
    & \multicolumn{1}{c|}{\textbf{56.8}}
    & \multicolumn{1}{c}{52.3}\\ 

    \multicolumn{1}{c|}{Qwen-VL-Chat \cite{bai2023qwen}} 
    & \multicolumn{1}{c}{47.0} 
    & \multicolumn{1}{c}{56.0}
    & \multicolumn{1}{c|}{55.1}
    & \multicolumn{1}{c}{52.7}\\ 

    \midrule

    % \multicolumn{1}{c}{\textit{\textbf{Our Training}}} 
    % & \multicolumn{1}{c}{} 
    % & \multicolumn{1}{c}{} 
    % & \multicolumn{1}{c}{}
    % & \multicolumn{1}{c}{}\\
    % \midrule
           
    % \multicolumn{1}{c|}{MIMO (w/o Aligner)} 
    % & \multicolumn{1}{c}{--} 
    % & \multicolumn{1}{c}{--}
    % & \multicolumn{1}{c|}{--}
    % & \multicolumn{1}{c}{--}\\ 

    \rowcolor{gray!10}
    \multicolumn{1}{c|}{MIMO} 
    & \multicolumn{1}{c}{\textbf{58.8}} 
    & \multicolumn{1}{c}{\textbf{57.0}}
    & \multicolumn{1}{c|}{52.4}
    & \multicolumn{1}{c}{\textbf{56.1}}\\ 
    \bottomrule
    \end{tabular}
    \label{tab:held-out-cap}
    \vspace{-0.5cm}
\end{table}

\subsection{Qualitative Analysis}

To compare the capabilities of MIMO with existing medical vision language models, we conduct case studies on questions from different perspectives. Figure \ref{fig:case}(a) shows the recognition and segmentation results of the model for organs in the bounding box under perspective II. It can be seen that SAM-Med can perform segmentation, but it cannot achieve the correspondence between the segmentation results and the labels. LLaVA-Med can answer the question, but due to the lack of visual association, the organ recognition results are wrong. MIMO can give segmentation results while answering the question, and has higher recognition accuracy. Figure \ref{fig:case}(b) shows the reasoning of the model on organ function under perspective III. LLaVA-Med only answers partial organs, while MIMO answers all the correct organs and gives the corresponding segmentation results. Figure \ref{fig:case}(c) shows the knowledge question answering results with the visual prompt under perspective IV. HuatuoGPT-Vision and MIMO both answer based on the internal knowledge of the model. However, MIMO can more intuitively give the corresponding segmentation results of the key medical entities in the answer.

\section{Conclusion}
\label{sec:conclusion}
In this paper, we present MIMO, the first medical multimodal large language model capable of simultaneously accepting visual prompts and generating text responses intertwined with grounded segmentation masks. To advance research and model development, we constructed MIMOSeg dataset, a large-scale, medical multimodal referring and grounding dataset, consisting of 895K samples. 
% Our data construction pipeline ensures both reliability and scalability. 
% Extensive experiments demonstrate that after training on MIMOSeg, MIMO performs well on both medical VQA and segmentation tasks.

\textbf{Limitations.} For the Q\&A generation approach used in constructing MIMOSeg, it is capable of generating the questions and answers only on the basis of the knowledge base. Consequently, its scope is bound by the given knowledge, limiting comprehensive coverage of information available for data generation. 

\newpage

\section*{Acknowledgment} 

This paper was supported by National Key R\&D Program of China (No. 2023YFC3502902, 2021YFF1201100). We thank Huamin Zhang's team from Institute of Basic Theory of Chinese Medicine, China Academy of Chinese Medical Sciences for assistance.
% \input{sec/3_finalcopy}

% \newpage
{
    \small
    \bibliographystyle{ieeenat_fullname}
    \bibliography{main}
}

% WARNING: do not forget to delete the supplementary pages from your submission 
\clearpage
\setcounter{page}{1}

\maketitlesupplementary

\appendix
\vspace{-2cm}
\section{Functional Comparison with Other Models}
\label{appendix:functional_compare}

\vspace{-0.2cm}
As is shown in Table \ref{tab:multimodal-comparison}. Current medical multimodal large language models \cite{li2023llava, alayrac2022flamingo,chen2024huatuogpt} face challenges in comprehensively supporting both referring and grounding capabilities. Meanwhile, some models in general domains are designed to address individual functions, including leveraging visual prompts to enhance referring capabilities \cite{wang2023caption,yuan2023osprey,chen2023shikra,cai2023making,lin2024draw} or generating mask segmentation outputs \cite{xia2023gsva,lai2024lisa,wei2024lasagna}. A line of works attempt to combine visual prompt inputs with mask segmentation outputs but lack grounding capabilities \cite{zhang2023next,zhang2024psalm}. Models \cite{you2023ferret, tian2024chatterbox} that incorporate both referring and grounding often rely on bounding box-based localization, which may be insufficient for high-precision requirements in medical contexts. A few models \cite{zhang2024groundhog,rasheed2024glamm} have shown significant advancements in integrating referring and pixel-wise grounding. However, these approaches often rely on complex visual prompt encoding mechanisms and extensive general-domain datasets, making direct adaptation to the medical field challenging.
Among these, our proposed MIMO adopts a concise design that integrates visual referring and pixel-level grounding capabilities. By supporting multimodal inputs and multimodal outputs, it offers an efficient and precise solution tailored to the unique demands of medical applications. 

\begin{table*}[t]
\fontsize{8}{11}\selectfont
\centering
\begin{tabular}{lccc|cccc|c}
\toprule
\multirow{2}{*}{\textbf{Method}} & \multicolumn{3}{c}{\textbf{Multimodal Input}} & \multicolumn{4}{c}{\textbf{Multimodal Output}} & \multirow{2}{*}{\makecell[c]{\textbf{Medical} \\ \textbf{LLM?}}} \\
\cmidrule[\cmidrulewidth](r){2-4} % 第一条横线，调整范围
\cmidrule[\cmidrulewidth](l){5-8} % 第二条横线，调整范围
& \multicolumn{1}{c}{\textbf{Image}} 
& \multicolumn{1}{c}{\textbf{Text}} 
& \multicolumn{1}{c}{\textbf{Visual Prompt}} 
& \multicolumn{1}{c}{\textbf{Text}} 
& \multicolumn{1}{c}{\textbf{Mask}} 
& \multicolumn{1}{c}{\textbf{\makecell[c]{\textbf{Mask-to-Phrase} \\ \textbf{Grounding}}}} 
& \multicolumn{1}{c}{\textbf{Multi-Region}} \\
\midrule
\rowcolor{magenta!3}
\multicolumn{1}{c}{LLaVA \cite{liu2024visual}} & \multicolumn{1}{c}{\color[HTML]{009F6B}\CheckmarkBold} & \multicolumn{1}{c}{\color[HTML]{009F6B}\CheckmarkBold} & \multicolumn{1}{c}{\color{red}\XSolidBrush} & \multicolumn{1}{c}{\color[HTML]{009F6B}\CheckmarkBold} & \multicolumn{1}{c}{\color{red}\XSolidBrush} & \multicolumn{1}{c}{\color{red}\XSolidBrush} & \multicolumn{1}{c}{\color{red}\XSolidBrush} & \multicolumn{1}{c}{\color{red}\XSolidBrush} \\

\multicolumn{1}{c}{miniGPT4 \cite{chen2023minigpt}} & \multicolumn{1}{c}{\color[HTML]{009F6B}\CheckmarkBold} & \multicolumn{1}{c}{\color[HTML]{009F6B}\CheckmarkBold} & \multicolumn{1}{c}{\color{red}\XSolidBrush} & \multicolumn{1}{c}{\color[HTML]{009F6B}\CheckmarkBold} & \multicolumn{1}{c}{\color{red}\XSolidBrush} & \multicolumn{1}{c}{\color{red}\XSolidBrush} & \multicolumn{1}{c}{\color{red}\XSolidBrush} & \multicolumn{1}{c}{\color{red}\XSolidBrush} \\
\rowcolor{magenta!3}
\multicolumn{1}{c}{mPLUG-OWL \cite{ye2023mplug}} & \multicolumn{1}{c}{\color[HTML]{009F6B}\CheckmarkBold} & \multicolumn{1}{c}{\color[HTML]{009F6B}\CheckmarkBold} & \multicolumn{1}{c}{\color{red}\XSolidBrush} & \multicolumn{1}{c}{\color[HTML]{009F6B}\CheckmarkBold} & \multicolumn{1}{c}{\color{red}\XSolidBrush} & \multicolumn{1}{c}{\color{red}\XSolidBrush} & \multicolumn{1}{c}{\color{red}\XSolidBrush} & \multicolumn{1}{c}{\color{red}\XSolidBrush} \\

\multicolumn{1}{c}{LLaMA-Adapter v2 \cite{gao2023llama}}  & \multicolumn{1}{c}{\color[HTML]{009F6B}\CheckmarkBold} & \multicolumn{1}{c}{\color[HTML]{009F6B}\CheckmarkBold} & \multicolumn{1}{c}{\color{red}\XSolidBrush} & \multicolumn{1}{c}{\color[HTML]{009F6B}\CheckmarkBold} & \multicolumn{1}{c}{\color{red}\XSolidBrush} & \multicolumn{1}{c}{\color{red}\XSolidBrush} & \multicolumn{1}{c}{\color{red}\XSolidBrush} & \multicolumn{1}{c}{\color{red}\XSolidBrush} \\
\rowcolor{magenta!3}
\multicolumn{1}{c}{InstructBLIP \cite{dai2023instructblip}} & \multicolumn{1}{c}{\color[HTML]{009F6B}\CheckmarkBold} & \multicolumn{1}{c}{\color[HTML]{009F6B}\CheckmarkBold} & \multicolumn{1}{c}{\color{red}\XSolidBrush} & \multicolumn{1}{c}{\color[HTML]{009F6B}\CheckmarkBold} & \multicolumn{1}{c}{\color{red}\XSolidBrush} & \multicolumn{1}{c}{\color{red}\XSolidBrush} & \multicolumn{1}{c}{\color{red}\XSolidBrush} & \multicolumn{1}{c}{\color{red}\XSolidBrush} \\

\multicolumn{1}{c}{LLaVA-med \cite{li2023llava}} & \multicolumn{1}{c}{\color[HTML]{009F6B}\CheckmarkBold} & \multicolumn{1}{c}{\color[HTML]{009F6B}\CheckmarkBold} & \multicolumn{1}{c}{\color{red}\XSolidBrush} & \multicolumn{1}{c}{\color[HTML]{009F6B}\CheckmarkBold} & \multicolumn{1}{c}{\color{red}\XSolidBrush} & \multicolumn{1}{c}{\color{red}\XSolidBrush} & \multicolumn{1}{c}{\color{red}\XSolidBrush} & \multicolumn{1}{c}{\color[HTML]{009F6B}\CheckmarkBold} \\
\rowcolor{magenta!3}
\multicolumn{1}{c}{HuatuoVision \cite{chen2024huatuogpt}} & \multicolumn{1}{c}{\color[HTML]{009F6B}\CheckmarkBold} & \multicolumn{1}{c}{\color[HTML]{009F6B}\CheckmarkBold} & \multicolumn{1}{c}{\color{red}\XSolidBrush} & \multicolumn{1}{c}{\color[HTML]{009F6B}\CheckmarkBold} & \multicolumn{1}{c}{\color{red}\XSolidBrush} & \multicolumn{1}{c}{\color{red}\XSolidBrush} & \multicolumn{1}{c}{\color{red}\XSolidBrush} & \multicolumn{1}{c}{\color[HTML]{009F6B}\CheckmarkBold} \\

\multicolumn{1}{c}{Med-Flamingo \cite{alayrac2022flamingo}} & \multicolumn{1}{c}{\color[HTML]{009F6B}\CheckmarkBold} & \multicolumn{1}{c}{\color[HTML]{009F6B}\CheckmarkBold} & \multicolumn{1}{c}{\color{red}\XSolidBrush} & \multicolumn{1}{c}{\color[HTML]{009F6B}\CheckmarkBold} & \multicolumn{1}{c}{\color{red}\XSolidBrush} & \multicolumn{1}{c}{\color{red}\XSolidBrush} & \multicolumn{1}{c}{\color{red}\XSolidBrush} & \multicolumn{1}{c}{\color[HTML]{009F6B}\CheckmarkBold} \\

\multicolumn{1}{c}{Caption Anything \cite{wang2023caption}} & \multicolumn{1}{c}{\color[HTML]{009F6B}\CheckmarkBold} & \multicolumn{1}{c}{\color[HTML]{009F6B}\CheckmarkBold} & \multicolumn{1}{c}{\color[HTML]{009F6B}\CheckmarkBold} & \multicolumn{1}{c}{\color[HTML]{009F6B}\CheckmarkBold} & \multicolumn{1}{c}{\color{red}\XSolidBrush} & \multicolumn{1}{c}{\color{red}\XSolidBrush} & \multicolumn{1}{c}{\color{red}\XSolidBrush} & \multicolumn{1}{c}{\color{red}\XSolidBrush} \\
\rowcolor{magenta!3}

\multicolumn{1}{c}{Osprey \cite{yuan2023osprey}} & \multicolumn{1}{c}{\color[HTML]{009F6B}\CheckmarkBold} & \multicolumn{1}{c}{\color[HTML]{009F6B}\CheckmarkBold} & \multicolumn{1}{c}{\color[HTML]{009F6B}\CheckmarkBold} & \multicolumn{1}{c}{\color[HTML]{009F6B}\CheckmarkBold} & \multicolumn{1}{c}{\color{red}\XSolidBrush} & \multicolumn{1}{c}{\color{red}\XSolidBrush} & \multicolumn{1}{c}{\color{red}\XSolidBrush} & \multicolumn{1}{c}{\color{red}\XSolidBrush} \\

\multicolumn{1}{c}{ViP-LLaVA \cite{cai2023making}} & \multicolumn{1}{c}{\color[HTML]{009F6B}\CheckmarkBold} & \multicolumn{1}{c}{\color[HTML]{009F6B}\CheckmarkBold} & \multicolumn{1}{c}{\color[HTML]{009F6B}\CheckmarkBold} & \multicolumn{1}{c}{\color[HTML]{009F6B}\CheckmarkBold} & \multicolumn{1}{c}{\color{red}\XSolidBrush} & \multicolumn{1}{c}{\color{red}\XSolidBrush} & \multicolumn{1}{c}{\color{red}\XSolidBrush} & \multicolumn{1}{c}{\color{red}\XSolidBrush} \\
\rowcolor{magenta!3}

\multicolumn{1}{c}{SPHINX-V \cite{lin2024draw}} & \multicolumn{1}{c}{\color[HTML]{009F6B}\CheckmarkBold} & \multicolumn{1}{c}{\color[HTML]{009F6B}\CheckmarkBold} & \multicolumn{1}{c}{\color[HTML]{009F6B}\CheckmarkBold} & \multicolumn{1}{c}{\color[HTML]{009F6B}\CheckmarkBold} & \multicolumn{1}{c}{\color{red}\XSolidBrush} & \multicolumn{1}{c}{\color{red}\XSolidBrush} & \multicolumn{1}{c}{\color{red}\XSolidBrush} & \multicolumn{1}{c}{\color{red}\XSolidBrush} \\

\multicolumn{1}{c}{Shikra \cite{chen2023shikra}} & \multicolumn{1}{c}{\color[HTML]{009F6B}\CheckmarkBold} & \multicolumn{1}{c}{\color[HTML]{009F6B}\CheckmarkBold} & \multicolumn{1}{c}{\color[HTML]{009F6B}\CheckmarkBold} & \multicolumn{1}{c}{\color[HTML]{009F6B}\CheckmarkBold} & \multicolumn{1}{c}{\color{red}\XSolidBrush} & \multicolumn{1}{c}{\color{red}\XSolidBrush} & \multicolumn{1}{c}{\color[HTML]{009F6B}\CheckmarkBold} & \multicolumn{1}{c}{\color{red}\XSolidBrush} \\
\rowcolor{magenta!3}
\multicolumn{1}{c}{ASMv2 \cite{wang2025all}} & \multicolumn{1}{c}{\color[HTML]{009F6B}\CheckmarkBold} & \multicolumn{1}{c}{\color[HTML]{009F6B}\CheckmarkBold} & \multicolumn{1}{c}{\color{red}\XSolidBrush} & \multicolumn{1}{c}{\color[HTML]{009F6B}\CheckmarkBold} & \multicolumn{1}{c}{\color{red}\XSolidBrush} & \multicolumn{1}{c}{\color[HTML]{009F6B}\CheckmarkBold \color{black}\textbf{/} \color{red}\XSolidBrush} & \multicolumn{1}{c}{\color[HTML]{009F6B}\CheckmarkBold} & \multicolumn{1}{c}{\color{red}\XSolidBrush} \\

\multicolumn{1}{c}{GSVA \cite{xia2024gsva}} & \multicolumn{1}{c}{\color[HTML]{009F6B}\CheckmarkBold} & \multicolumn{1}{c}{\color[HTML]{009F6B}\CheckmarkBold} & \multicolumn{1}{c}{\color{red}\XSolidBrush} & \multicolumn{1}{c}{\color[HTML]{009F6B}\CheckmarkBold} & \multicolumn{1}{c}{\color[HTML]{009F6B}\CheckmarkBold} & \multicolumn{1}{c}{\color{red}\XSolidBrush} & \multicolumn{1}{c}{\color[HTML]{009F6B}\CheckmarkBold} & \multicolumn{1}{c}{\color{red}\XSolidBrush} \\
\rowcolor{magenta!3}

\multicolumn{1}{c}{LaSagnA \cite{wei2024lasagna}} & \multicolumn{1}{c}{\color[HTML]{009F6B}\CheckmarkBold} & \multicolumn{1}{c}{\color[HTML]{009F6B}\CheckmarkBold} & \multicolumn{1}{c}{\color{red}\XSolidBrush} & \multicolumn{1}{c}{\color[HTML]{009F6B}\CheckmarkBold} & \multicolumn{1}{c}{\color[HTML]{009F6B}\CheckmarkBold} & \multicolumn{1}{c}{\color[HTML]{009F6B}\CheckmarkBold} & \multicolumn{1}{c}{\color[HTML]{009F6B}\CheckmarkBold} & \multicolumn{1}{c}{\color{red}\XSolidBrush} \\

\multicolumn{1}{c}{LISA \cite{lai2024lisa}} & \multicolumn{1}{c}{\color[HTML]{009F6B}\CheckmarkBold} & \multicolumn{1}{c}{\color[HTML]{009F6B}\CheckmarkBold} & \multicolumn{1}{c}{\color{red}\XSolidBrush} & \multicolumn{1}{c}{\color[HTML]{009F6B}\CheckmarkBold} & \multicolumn{1}{c}{\color[HTML]{009F6B}\CheckmarkBold} & \multicolumn{1}{c}{\color{red}\XSolidBrush} & \multicolumn{1}{c}{\color{red}\XSolidBrush} & \multicolumn{1}{c}{\color{red}\XSolidBrush} \\
\rowcolor{magenta!3}

\multicolumn{1}{c}{PSALM \cite{zhang2024psalm}} & \multicolumn{1}{c}{\color[HTML]{009F6B}\CheckmarkBold} & \multicolumn{1}{c}{\color[HTML]{009F6B}\CheckmarkBold} & \multicolumn{1}{c}{\color[HTML]{009F6B}\CheckmarkBold} & \multicolumn{1}{c}{\color[HTML]{009F6B}\CheckmarkBold} & \multicolumn{1}{c}{\color[HTML]{009F6B}\CheckmarkBold} & \multicolumn{1}{c}{\color{red}\XSolidBrush} & \multicolumn{1}{c}{\color{red}\XSolidBrush}  & \multicolumn{1}{c}{\color{red}\XSolidBrush} \\

\multicolumn{1}{c}{NextChat \cite{zhang2023next}} & \multicolumn{1}{c}{\color[HTML]{009F6B}\CheckmarkBold} & \multicolumn{1}{c}{\color[HTML]{009F6B}\CheckmarkBold} & \multicolumn{1}{c}{\color[HTML]{009F6B}\CheckmarkBold} & \multicolumn{1}{c}{\color[HTML]{009F6B}\CheckmarkBold} & \multicolumn{1}{c}{\color[HTML]{009F6B}\CheckmarkBold} & \multicolumn{1}{c}{\color{red}\XSolidBrush} & \multicolumn{1}{c}{\color[HTML]{009F6B}\CheckmarkBold}  & \multicolumn{1}{c}{\color{red}\XSolidBrush} \\
\rowcolor{magenta!3}

\multicolumn{1}{c}{Ferret \cite{you2023ferret}} & \multicolumn{1}{c}{\color[HTML]{009F6B}\CheckmarkBold} & \multicolumn{1}{c}{\color[HTML]{009F6B}\CheckmarkBold} & \multicolumn{1}{c}{\color[HTML]{009F6B}\CheckmarkBold} & \multicolumn{1}{c}{\color[HTML]{009F6B}\CheckmarkBold} & \multicolumn{1}{c}{\color{red}\XSolidBrush} & \multicolumn{1}{c}{\color[HTML]{009F6B}\CheckmarkBold \color{black}\textbf{/} \color{red}\XSolidBrush} & \multicolumn{1}{c}{\color[HTML]{009F6B}\CheckmarkBold} & \multicolumn{1}{c}{\color{red}\XSolidBrush} \\

\multicolumn{1}{c}{ChatterBox \cite{tian2024chatterbox}} & \multicolumn{1}{c}{\color[HTML]{009F6B}\CheckmarkBold} & \multicolumn{1}{c}{\color[HTML]{009F6B}\CheckmarkBold} & \multicolumn{1}{c}{\color[HTML]{009F6B}\CheckmarkBold} & \multicolumn{1}{c}{\color[HTML]{009F6B}\CheckmarkBold} & \multicolumn{1}{c}{\color{red}\XSolidBrush} & \multicolumn{1}{c}{\color[HTML]{009F6B}\CheckmarkBold \color{black}\textbf{/} \color{red}\XSolidBrush} & \multicolumn{1}{c}{\color{red}\XSolidBrush}  & \multicolumn{1}{c}{\color{red}\XSolidBrush} \\
\rowcolor{magenta!3}

\multicolumn{1}{c}{GroundHog \cite{zhang2024groundhog}} & \multicolumn{1}{c}{\color[HTML]{009F6B}\CheckmarkBold} & \multicolumn{1}{c}{\color[HTML]{009F6B}\CheckmarkBold} & \multicolumn{1}{c}{\color[HTML]{009F6B}\CheckmarkBold$^*$} & \multicolumn{1}{c}{\color[HTML]{009F6B}\CheckmarkBold} & \multicolumn{1}{c}{\color[HTML]{009F6B}\CheckmarkBold} & \multicolumn{1}{c}{\color[HTML]{009F6B}\CheckmarkBold} & \multicolumn{1}{c}{\color[HTML]{009F6B}\CheckmarkBold} & \multicolumn{1}{c}{\color{red}\XSolidBrush} \\

\multicolumn{1}{c}{GLaMM \cite{rasheed2024glamm}} & \multicolumn{1}{c}{\color[HTML]{009F6B}\CheckmarkBold} & \multicolumn{1}{c}{\color[HTML]{009F6B}\CheckmarkBold} & \multicolumn{1}{c}{\color[HTML]{009F6B}\CheckmarkBold$^*$} & \multicolumn{1}{c}{\color[HTML]{009F6B}\CheckmarkBold} & \multicolumn{1}{c}{\color[HTML]{009F6B}\CheckmarkBold} & \multicolumn{1}{c}{\color[HTML]{009F6B}\CheckmarkBold} & \multicolumn{1}{c}{\color[HTML]{009F6B}\CheckmarkBold} & \multicolumn{1}{c}{\color{red}\XSolidBrush} \\

\rowcolor{magenta!18}
\multicolumn{1}{c}{MIMO(ours)} & \multicolumn{1}{c}{\color[HTML]{009F6B}\CheckmarkBold} & \multicolumn{1}{c}{\color[HTML]{009F6B}\CheckmarkBold} & \multicolumn{1}{c}{\color[HTML]{009F6B}\CheckmarkBold} & \multicolumn{1}{c}{\color[HTML]{009F6B}\CheckmarkBold} & \multicolumn{1}{c}{\color[HTML]{009F6B}\CheckmarkBold} & \multicolumn{1}{c}{\color[HTML]{009F6B}\CheckmarkBold} & \multicolumn{1}{c}{\color[HTML]{009F6B}\CheckmarkBold} & \multicolumn{1}{c}{\color[HTML]{009F6B}\CheckmarkBold} \\
\bottomrule
\end{tabular}
\caption{Comparison of recent multimodal large language models. {\color[HTML]{009F6B}\CheckmarkBold} indicates support, while {\color{red}\XSolidBrush} indicates no support.\quad {\color[HTML]{009F6B}\CheckmarkBold \color{black}\textbf{/} \color{red}\XSolidBrush} ASMv2 does not support pixel-wise segmentation masks with phrase grounding, using bounding boxes instead, making it unsuitable for medical applications requiring fine-grained precision. \quad {\color[HTML]{009F6B}\CheckmarkBold \color{black}\textbf{/} \color{red}\XSolidBrush} Ferret and ChatterBox output bounding boxes for localization instead of masks and does not support pixel-wise grounding.\quad {\color[HTML]{009F6B}\CheckmarkBold$^*$} In GLaMM, user-input spatial prompts are limited to bounding boxes, with a carefully designed region encoder extracting a region-of-interest representation corresponding to the box. \quad {\color[HTML]{009F6B}\CheckmarkBold$^*$} GroundHog converts user-input spatial prompts into binary masks via SAM and uses a masked feature extractor to extract local features. }
\label{tab:multimodal-comparison}
\end{table*}

\vspace{-0.2cm}
\section{Data Statistics}
\label{appendix:data_statistics}
\vspace{-0.1cm}
\subsection{Data Modalities}
\label{appendix:data_modalities}
\begin{figure*}[htbp]
  \centering
  \includegraphics[width=0.8\textwidth]{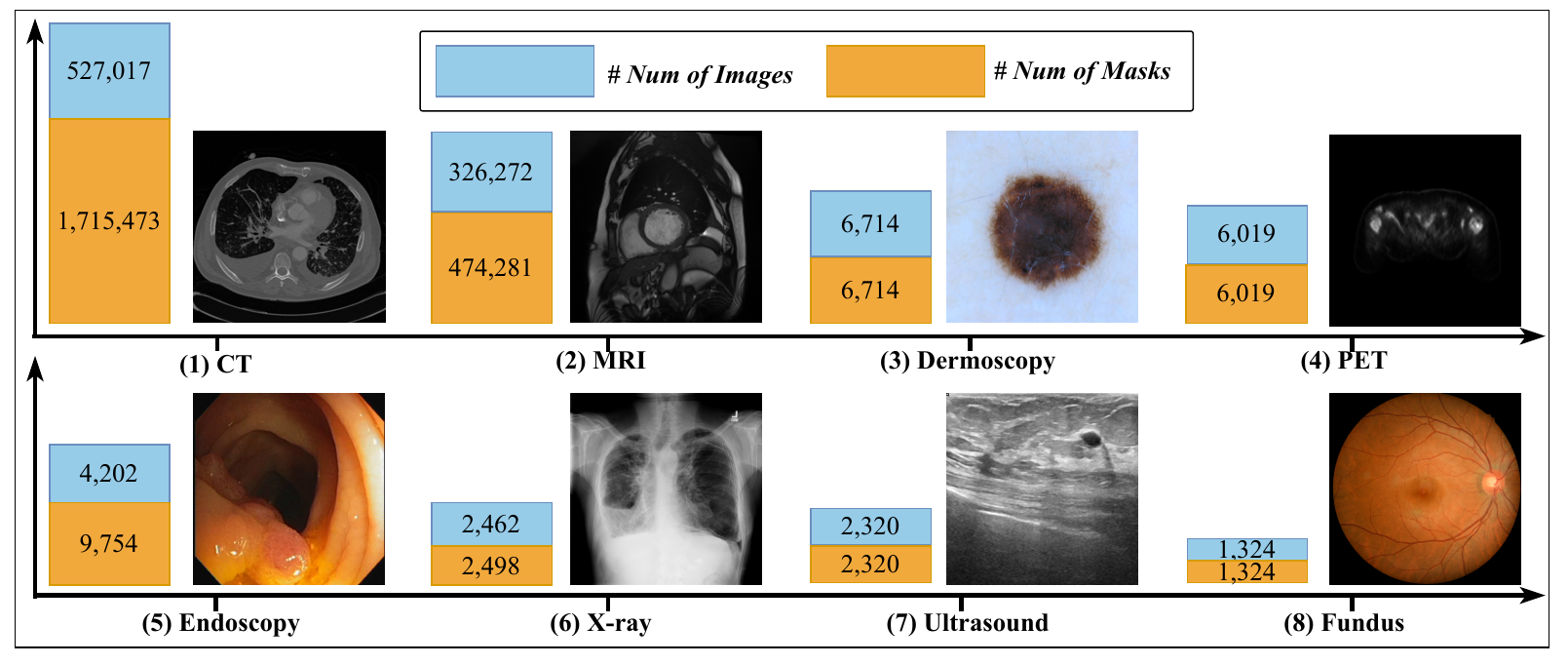} 
  \caption{Modal statistics results of MIMOSeg.}
  \label{fig:data_mod}
  \vspace{-0.5cm}
\end{figure*}

\vspace{-0.2cm}
Figure \ref{fig:data_mod} shows the modality statistics of MIMOSeg. MIMOSeg contains eight types of images of different medical modalities. The order from most to least is CT, MRI, Dermoscopy, PET, Endoscopy, X-Ray, Ultrasound and Fundus. %{\color{magenta}{Our dataset will be publicly released.}}

\vspace{-0.2cm}
\subsection{Data Source}
\label{appendix:data_source}
\vspace{-0.2cm}
Table \ref{tab:datasource} shows the data source of MIMOSeg. For each data source, we count the number of images, the number of masks, the modality type of the dataset and the labels of the dataset, and show which perspective the dataset is used to construct data.

\vspace{-0.25cm}
\subsection{Statistics of Different Perspectives}
\label{appendix:statistics_perspectives}
\vspace{-0.2cm}
Table \ref{tab:ds_for_mimoseg} shows the statistical results of MIMOSeg, including the number of training, validation, and test sets for each perspective. Since the amount of MIMOSeg data is very large, we divide the training, validation, and test sets in a ratio of 99:0.5:0.5.

\begin{table}[htbp]
    \centering
    \setlength{\tabcolsep}{5pt}
    \caption{Data Statistics for MIMOSeg, including different perspectives.}
    \label{tab:ds_for_mimoseg}
      % \captionsetup{justification=centering}
    \fontsize{8}{10}\selectfont    %{字体尺寸}{行距}
    \begin{tabular}{l|ccc|c}
        \toprule
         \textbf{Dataset} & \textbf{Train} & \textbf{Val} & \textbf{Test} &  \textbf{Total} \\
         \midrule
         \textbf{Perspective I} & 249,665 & 2,564 & 2,545 & 254,774   \\
         \textbf{Perspective II} & 249,698 & 2,552 & 2,334 & 254,584  \\
         \textbf{Perspective III} & 181,046 & 726 & 726 & 182,498 \\
         \textbf{Perspective IV} & 178,594 & 1,070 & 1,070 & 180,734 \\
         \midrule
         \textbf{Total} & 859,003 & 6,912 & 6,675 & 872,590\\
        \bottomrule
    \end{tabular}
\end{table}

\subsection{Statistics of Zero-shot Datasets}

To demonstrate the generalization ability of the model, we conduct zero-shot tests on 6 held-out datasets. Table \ref{tab:ds_for_zero_shot} shows the statistical results of the number of each zero-shot dataset.

\begin{table}[htbp]
    \centering
    \setlength{\tabcolsep}{5pt}
    \caption{Data Statistics for held-out experiments.}
    \label{tab:ds_for_zero_shot}
      % \captionsetup{justification=centering}
    \fontsize{8}{10}\selectfont    %{字体尺寸}{行距}
    \begin{tabular}{cccc}
        \toprule
         \textbf{Segmentation} & \textbf{X-ray} & \textbf{Fundus} &
         \textbf{Skin Lesion}\\
         \midrule
         \textbf{Num} & 281 & 270 &
         206\\
         \midrule
         \textbf{VQA} & \textbf{VQA-RAD} & \textbf{SLAKE} &
         \textbf{PathVQA}\\
         \midrule
         \textbf{Num} & 272 & 416 &
         3,391\\
         
        \bottomrule
    \end{tabular}
\end{table}

\section{Data Analysis}
\label{appendix:data_analysis}

\begin{figure*}[htbp!]
  \centering
  \includegraphics[width=0.8\textwidth]{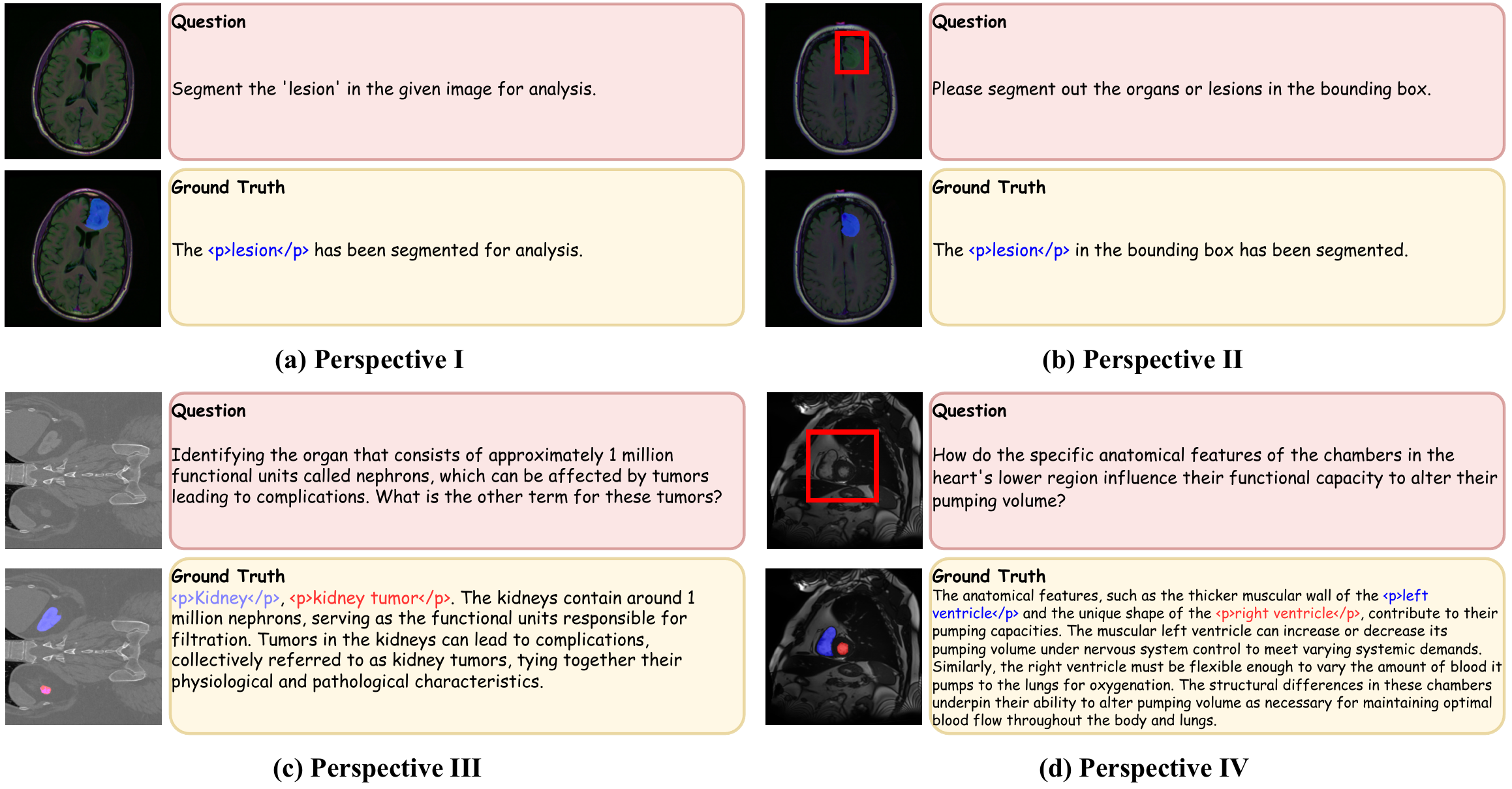} 
  \caption{Several examples of MIMOSeg.}
  \label{fig:data_examples}
  \vspace{-0.5cm}
\end{figure*}

Figure \ref{fig:data_examples} presents several examples of MIMOSeg. As mentioned, different perspectives focus on different aspects. Perspective I focuses on the model's ability to directly follow instructions. Perspective II focuses on the model's ability to understand visual clues. Perspective III addresses segmentation associated with complex reasoning. Perspective IV focuses on complex question answering with visual clues. In the divided test and validation sets, we further manually filtered the test and validation sets for Perspective III and Perspective IV to obtain high-quality evaluation data. Specially, we manually excluded data containing questions or answers unrelated to the image content.

\section{More Implementation Details}
\label{appendix:more_implementation_details}

\subsection{Evaluation metric}
\textbf{F1 Score.} Inspired by \cite{you2023ferret,zhou2019grounded}, we propose the F1 Score to quantify the grounding capability of masks aligned with medical entities. For the multimodel output $R=\left\{t_{j} \mid j=1,2, \ldots, n\right\} \nonumber 
\cup\left\{<c_{i},s_{i}> \mid s_i \in S, c_i \in T,i=1,2, \ldots, m\right\}$, we define $A$ as the total number of entity words ($c$) in the generated sentence $R$, and $B$ as the total number of entity words in the ground truth sentence. $E$ represents the total number of correct prediction pairs ($\{<c,s>\}$). In this paper, a correct prediction refers to generating entity words that match those in the ground truth and producing a correct mask segmentation (i.e., IoU with the ground truth segmentation $> 0.5$). In F1, the precision and recall can be defined as

\begin{equation}
\text{Precision} = \frac{E}{A} , \quad
\text{Recall} = \frac{E}{B}.
\end{equation}
The F1 score is calculated as follows:
\begin{equation}
F_1 = \frac{2 \cdot \text{Precision} \cdot \text{Recall}}{\text{Precision} + \text{Recall}}.
\end{equation}

\subsection{Model Architecture and Training}
We use Vicuna LLM with 7B parameters as the default large language model and instantiate the image encoder with ViT-H/14 CLIP. The visual prompt encoder adopts a positional encoding scheme similar to SAM. The multi-modal input aligner is randomly initialized, while the segmentation mask encoder and mask decoder are initialized using SAM's encoder and decoder, respectively. During training, the image encoder and segmentation mask encoder remain frozen, while the visual prompt encoder, multi-modal input aligner, and mask decoder are trained. Additionally, we apply Low-Rank Adaptation (LoRA) \cite{hu2021lora} with  $\alpha = 8$ to fine-tune the LLM. {\color{magenta}{Our codes and pretrained models will be publicly released.}}

After initialization, MIMO was trained using the aforementioned MIMOSeg dataset and the 60K LLaVA-Med VQA dataset on 4 A800 GPUs for 3 epochs, which took approximately 10$\sim$12 days. The training was optimized using the Adam optimizer with a learning rate of 3e-4 and a batch size of 40. To enhance the model's capability to follow VQA instructions, the trained model is further fine-tuned for one additional epoch on the LLaVA-Med VQA dataset, with the mask decoder frozen.

\section{Instructions and Prompts}
\label{appendix:instructions}

\subsection{Instruction Templates for Perspective I \& II}
\label{appendix:instructions_I_II}

\subsubsection{Instruction Templates for Perspective I}
To construct the instruction-following dataset in Perspective I, we design template instructions and responses. Specifically, for a given image, its corresponding masks, and the labels associated with the masks, we design segmentation instruction templates that specify the names of the medical entities to be segmented. Depending on the number of labels in each image, the responses use either single-label or multi-label response templates, as shown in Table \ref{tab:perspective1-prompt}.

\begin{table}[ht]
    \fontsize{8}{9}\selectfont    %{字体尺寸}{行距}
    \centering
    \resizebox{\linewidth}{!}{
    \begin{tabular}{|c|}
    \toprule
    \rowcolor{gray!30}
    \multicolumn{1}{l}{\textbf{Instruction Template of of Perspective I }}\\
    
    \multicolumn{1}{l}{\text{Please segment the \{\} in the image.}}\\
    \multicolumn{1}{l}{\text{Can you identify and segment the distinct \{\} elements within the image?}}\\
    \multicolumn{1}{l}{\text{I need the \{\} in the image to be categorized into individual segments.}}\\
    \multicolumn{1}{l}{\text{Could you analyze the image and segment the \{\} into separate segments?}}\\
    \multicolumn{1}{l}{\text{Can you perform an image segmentation to extract the \{\}?}}\\
    \multicolumn{1}{l}{\text{Segment the \{\} in the given image for analysis.}}\\
    \multicolumn{1}{l}{\text{Segment and highlight the \{\} in the image.}}\\
    \multicolumn{1}{l}{\text{Cut out the \{\} from the image and display it.}}\\
    \multicolumn{1}{l}{\text{Segment the \{\} regions in the medical image.}}\\

    \midrule
    
    \rowcolor{gray!30}
    \multicolumn{1}{l}{\textbf{Response Template of Perspective I for single-label image}}\\

    \multicolumn{1}{l}{\text{The image includes \{\}. The segmentation result is shown in the image.}}\\
    \multicolumn{1}{l}{\text{The segmentation result is displayed in the image, which includes \{\}.}}\\
    \multicolumn{1}{l}{\text{You can see the segmentation result in the image, along with \{\}.}}\\
    \multicolumn{1}{l}{\text{The image shows the segmentation result, which includes \{\}.}}\\
    \multicolumn{1}{l}{\text{The segmentation result in the image includes \{\}.}}\\
    \multicolumn{1}{l}{\text{Within the image, \{\} is present, and the segmentation result is visible.}}\\

    \rowcolor{gray!30}
    \multicolumn{1}{l}{\textbf{Response Template of Perspective I for multi-label image}}\\

    \multicolumn{1}{l}{\text{The image includes \{\}. The segmentation results are shown in the image.}}\\
    \multicolumn{1}{l}{\text{The segmentation results are displayed in the image, which include \{\}.}}\\
    \multicolumn{1}{l}{\text{You can see the segmentation results in the image, along with \{\}.}}\\
    \multicolumn{1}{l}{\text{The image shows the segmentation results, which include \{\}.}}\\
    \multicolumn{1}{l}{\text{The segmentation results in the image include \{\}.}}\\
    \multicolumn{1}{l}{\text{Within the image, \{\} are present, and the segmentation results are visible.}}\\

    \bottomrule
    \end{tabular}
    }
    \caption{Templates used for the Perspective I.} 
    \label{tab:perspective1-prompt}
\end{table}

\subsubsection{Instruction Templates for Perspective II}
In constructing the instruction-following dataset for Perspective II, we design question templates to trigger segmentation. Depending on the different forms of visual prompts, we formulate two types of question templates (i.e. box and point) along with corresponding answer templates, as shown in Table \ref{tab:perspective2-prompt}.

\begin{table}[ht]
    \fontsize{7}{9}\selectfont    %{字体尺寸}{行距}
    \centering
    \begin{tabular}{|c|}
    \toprule
    \rowcolor{gray!30}
    \multicolumn{1}{l}{\textbf{Instruction Template of of Perspective II for box prompt}}\\
    
   \multicolumn{1}{l}{\text{Please segment out the organs or lesions in the bounding box.}}\\
    \multicolumn{1}{l}{\text{Please identify and segment the organs or lesions within the given bounding box.}}\\
    \multicolumn{1}{l}{\text{Segment the organs or lesions that are located inside the bounding box.}}\\
    \multicolumn{1}{l}{\text{Can you segment out the organs or lesions found in the specified bounding box?}}\\
    \multicolumn{1}{l}{\text{Please perform segmentation of the organs or lesions within this bounding box.}}\\
    \multicolumn{1}{l}{\text{Segment any organs or lesions present within the provided bounding box.}}\\
    \multicolumn{1}{l}{\text{Conduct segmentation of organs or lesions contained in the bounding box.}}\\
    \multicolumn{1}{l}{\text{Could you segment the organs or lesions that are inside the bounding box?}}\\

    \midrule
    
    \rowcolor{gray!30}
    \multicolumn{1}{l}{\textbf{Response Template of Percpective II for box prompt}}\\

    \multicolumn{1}{l}{\text{The result of segmentation is \{\} and is shown in the image.}}\\
    \multicolumn{1}{l}{\text{The outcome of the segmentation is \{\} and is displayed in the image.}}\\
    \multicolumn{1}{l}{\text{The segmentation result is \{\} and is shown in the image.}}\\
    \multicolumn{1}{l}{\text{The organs or lesions have been segmented and the result is \{\}.}}\\
    \multicolumn{1}{l}{\text{The segmentation output is \{\} and is present in the image.}}\\
    \multicolumn{1}{l}{\text{Segmentation results in \{\}, which is shown in the image.}}\\

    \midrule
    
    \rowcolor{gray!30}
    \multicolumn{1}{l}{\textbf{Instruction Template of Perspective II for point prompt}}\\

    \multicolumn{1}{l}{\text{Please segment out the organs or lesions at the specified point.}}\\
    \multicolumn{1}{l}{\text{Please identify and segment the organs or lesions at the given point.}}\\
    \multicolumn{1}{l}{\text{Segment the organs or lesions that are located at the specified point.}}\\
    \multicolumn{1}{l}{\text{Can you segment out the organs or lesions found at the specified point?}}\\
    \multicolumn{1}{l}{\text{Please perform segmentation of the organs or lesions at this point.}}\\
    \multicolumn{1}{l}{\text{Segment any organs or lesions present at the provided point.}}\\
    \multicolumn{1}{l}{\text{Conduct segmentation of organs or lesions at the specified point.}}\\
    \multicolumn{1}{l}{\text{Could you segment the organs or lesions at the specified point?}}\\

    \midrule
    
    \rowcolor{gray!30}
    \multicolumn{1}{l}{\textbf{Response Template of Perspective II for point prompt}}\\

    \multicolumn{1}{l}{\text{The result of segmentation is \{\} and is shown in the image.}}\\
    \multicolumn{1}{l}{\text{The outcome of the segmentation is \{\} and is displayed in the image.}}\\
    \multicolumn{1}{l}{\text{The segmentation result is \{\} and is shown in the image.}}\\
    \multicolumn{1}{l}{\text{The organs or lesions have been segmented and the result is \{\}.}}\\
    \multicolumn{1}{l}{\text{The segmentation output is \{\} and is present in the image.}}\\
    \multicolumn{1}{l}{\text{Segmentation results in \{\}, which is shown in the image.}}\\

    \bottomrule
    \end{tabular}
    \caption{Templates used for the Perspective II.} 
    \label{tab:perspective2-prompt}
\end{table}

\subsection{Prompts for Q\&A generation in Perspective III \& IV}
\label{appendix:prompts_III_IV}

\subsubsection{Prompts for Q\&A generation in Perspective III}
\begin{figure*}[ht]
  \centering
  \includegraphics[width=0.8\linewidth]{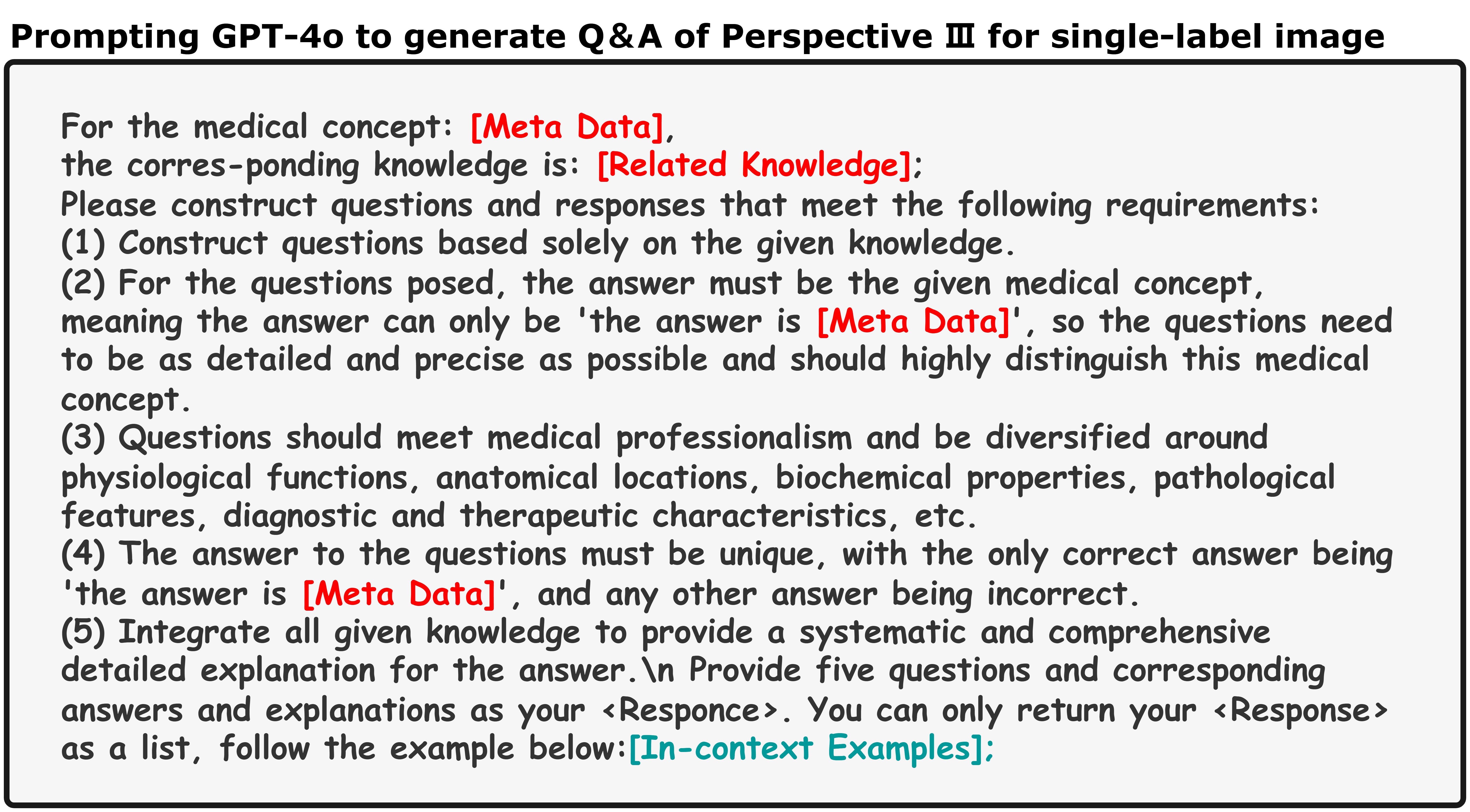} 
  \caption{Knowledge-based Prompt for Q\&A generation on single-label data in Perspective III.}
  \label{fig:prompt3-single}
  \vspace{0.5cm}
\end{figure*}

\begin{figure*}[ht]
  \centering
  \includegraphics[width=0.8\linewidth]{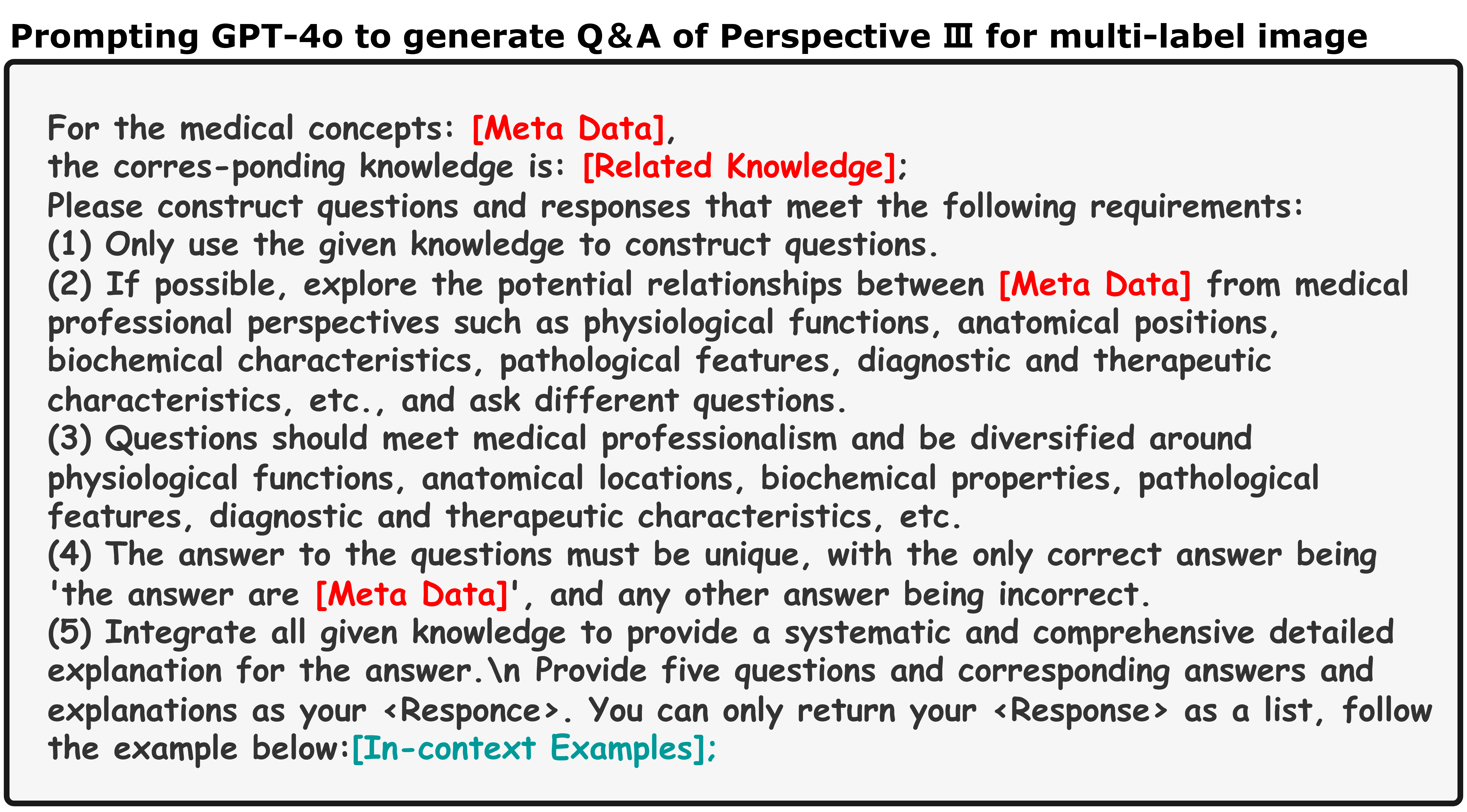} 
  \caption{Knowledge-based Prompt for Q\&A generation on multi-label data in Perspective III.}
  \label{fig:prompt3-multi}
  \vspace{0.5cm}
\end{figure*}

\begin{figure*}[ht]
  \centering
  \includegraphics[width=0.8\linewidth]{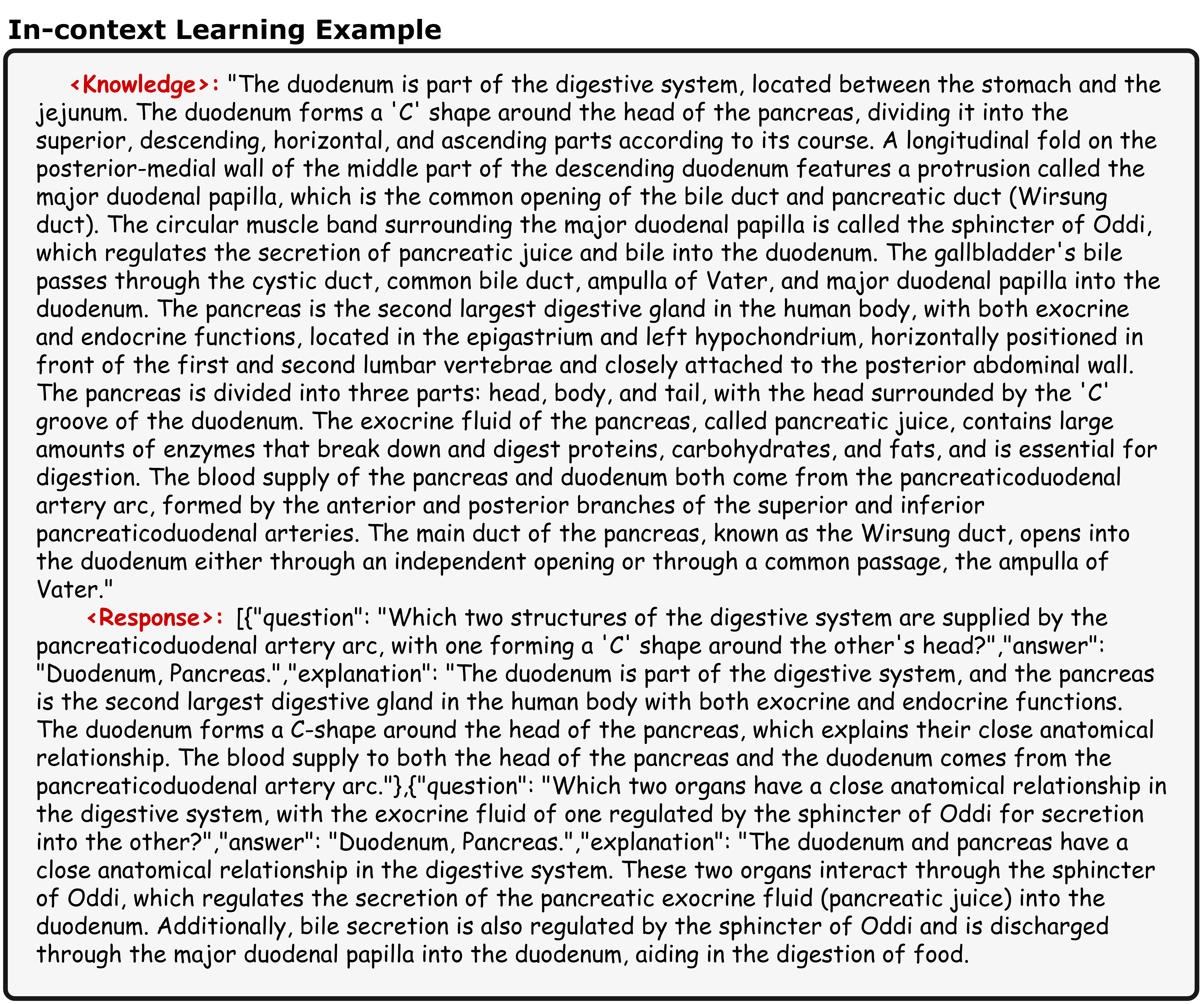} 
  \caption{An in-context learning example.}
  \label{fig:prompt3-incontext}
  \vspace{0.5cm}
\end{figure*}

Figure \ref{fig:prompt3-single} and Figure \ref{fig:prompt3-multi} illustrate the knowledge-based prompts used to construct the Perspective III data for MIMOSeg. Each prompt contains medical entity names, their corresponding knowledge, and in-context learning examples.  An in-context learning example is shown in Figure \ref{fig:prompt3-incontext}. For images with multiple mask labels, we concatenate the knowledge associated with the entity labels and input it into the prompt, emphasizing the relationships between multiple entities.

\subsubsection{Prompts for Q\&A generation in Perspective IV}
\begin{figure*}[ht]
  \centering
  \includegraphics[width=0.8\linewidth]{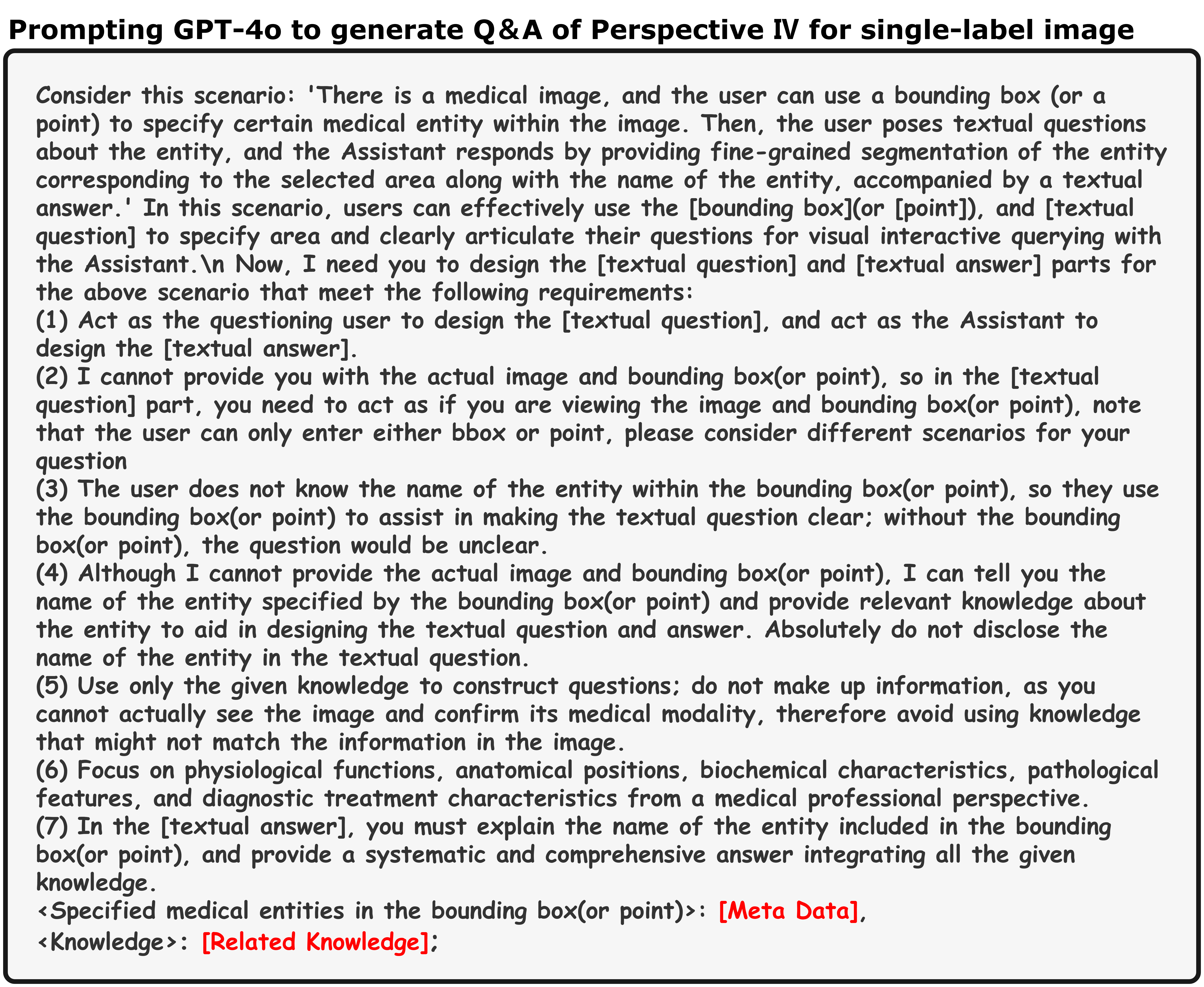} 
  \caption{Knowledge-based Prompt for Q\&A generation on single-label data in Perspective IV.}
  \label{fig:prompt4-single}
  \vspace{0.5cm}
\end{figure*}

\begin{figure*}[ht]
  \centering
  \includegraphics[width=0.8\linewidth]{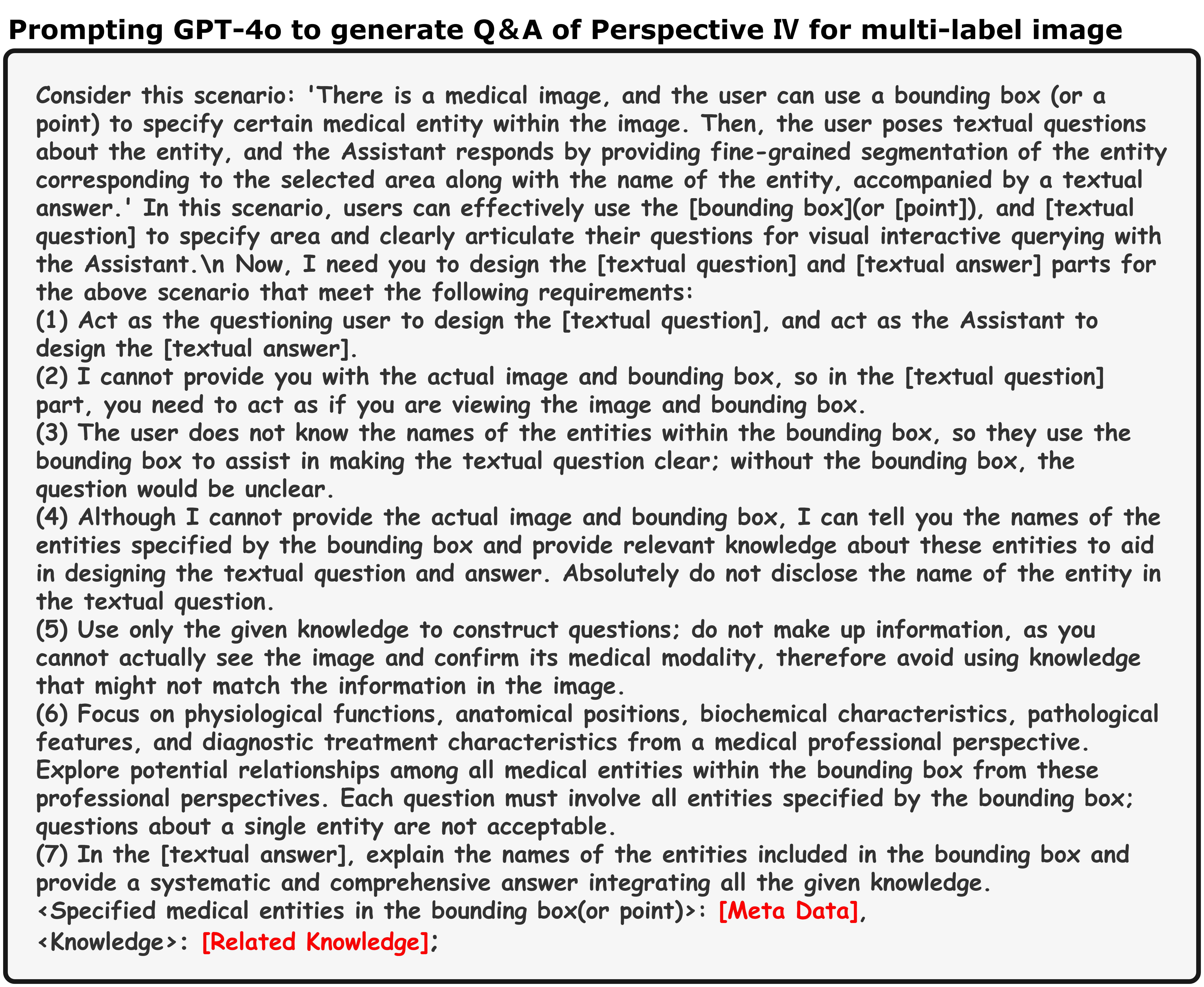} 
  \caption{Knowledge-based Prompt for Q\&A generation on multi-label data in Perspective IV.}
  \label{fig:prompt4-multi}
  \vspace{0.5cm}
\end{figure*}

Figure \ref{fig:prompt4-single} and Figure \ref{fig:prompt4-multi} present the knowledge-based prompts used to construct the Perspective IV data for MIMOSeg. Each prompt includes medical entity names and their corresponding knowledge. For images with multiple mask labels, we concatenate the knowledge associated with each entity label and input it into the prompt, focusing on the relationsHips between the entities.

\section{More Experimental Results}
\label{appendix:more_experimental_results}

\begin{figure*}[htbp!]
  \centering
  \includegraphics[width=0.8\textwidth]{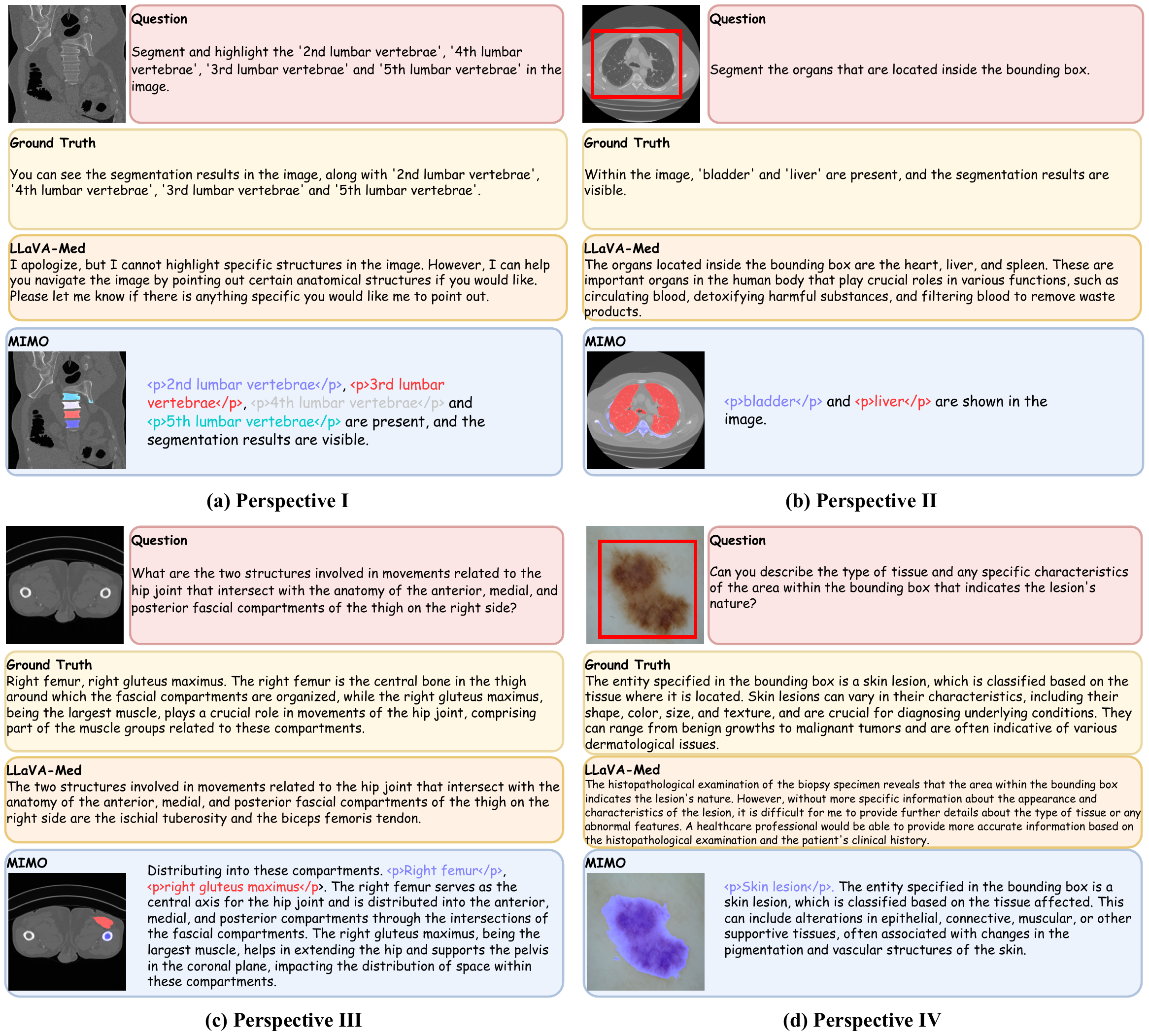} 
  \caption{Qualitative analysis compared with LLaVA-Med.}
  \label{fig:case_llavamed}
  \vspace{-0.5cm}
\end{figure*}

\begin{figure*}[htbp!]
  \centering
  \includegraphics[width=0.8\textwidth]{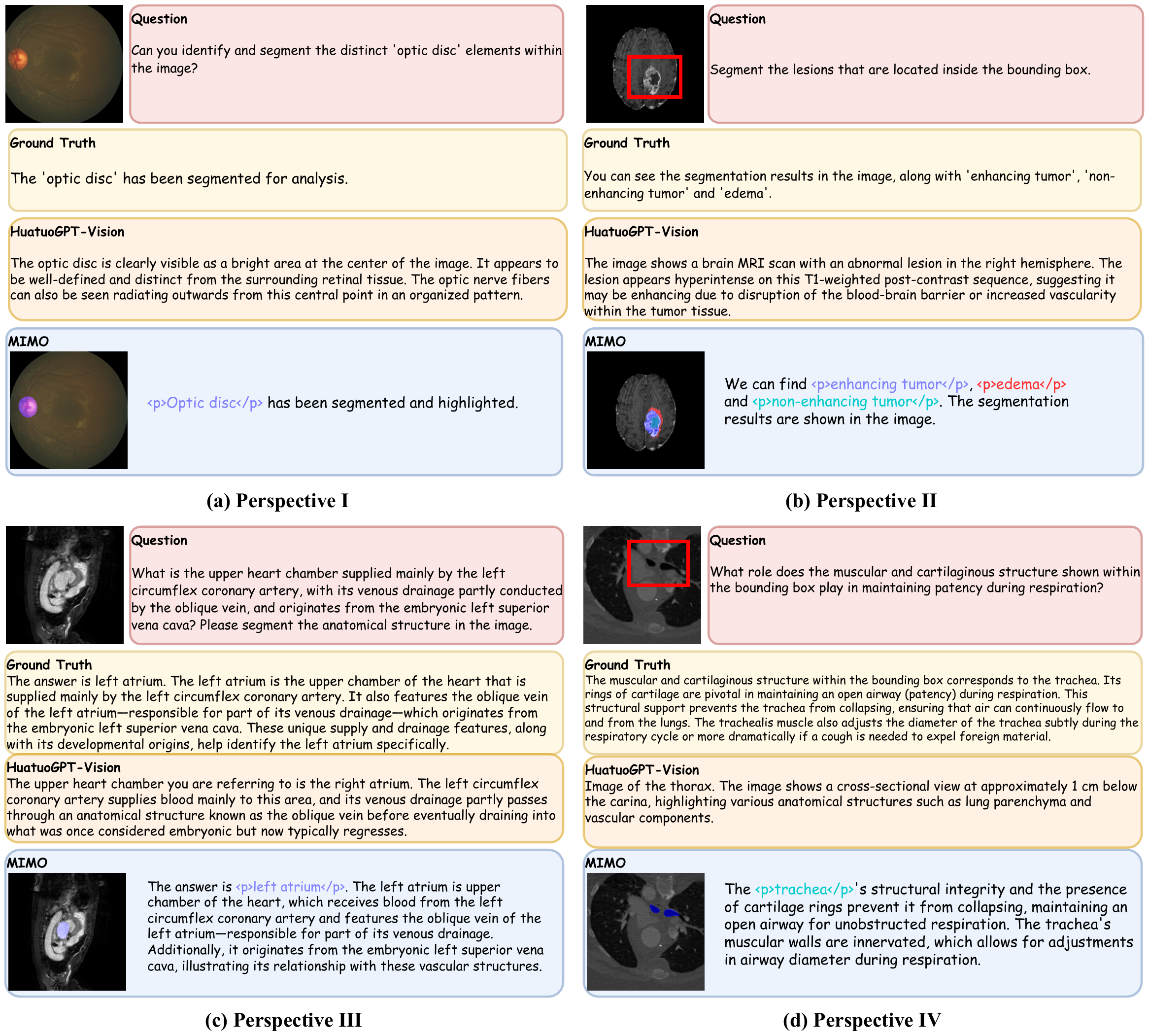} 
  \caption{Qualitative analysis compared with HuatuoGPT-Vision.}
  \label{fig:case_huatuo}
  \vspace{-0.5cm}
\end{figure*}

Figures \ref{fig:case_llavamed} and Figure \ref{fig:case_huatuo} show the case analysis results of LLaVA-Med and HuatuoGPT-Vision under four perspectives. MIMO can provide the segmentation results of the medical entities associated with the answer while giving the correct text reply.

% \section{Limitations and Future Work}
% \label{appendix:limitations}

\section{Ethics Statement}
\label{appendix:ethics}
MIMO is built upon a large language model (LLM), inheriting original limitations of LLMs, such as language hallucinations, where it may generate harmful or counterfactual responses. Moreover, machines are not infallible, and there is a potential risk that the model may misinterpret user inputs or make inaccurate predictions. In high-risk medical settings, such errors could be harmful or even dangerous. Researchers and developers must be aware of the potential risks associated with the use and misuse of medical LLMs in healthcare environments and implement both automated safeguards (e.g., setting strict thresholds for diagnostic suggestions) and human interventions (e.g., training staff to recognize potential system failures). We explicitly state that the MIMOSeg dataset we release is intended solely for research purposes. Furthermore, our data collection methods comply with the terms of use and adhere to the intellectual property and privacy rights of the original authors.

\linespread{1.1}
\begin{table*}[htbp!]
    \caption{Data source of MIMOSeg. P-I, P-II, P-III, and P-IV respectively represent whether the dataset is applied to the data construction of Perspective I, Perspective II, Perspective III, and Perspective IV.} 
    \fontsize{8}{8.5}\selectfont    %{字体尺寸}{行距}
    \tabcolsep=0.1cm %列间距
    \centering
    \begin{tabular}{c|rrrrrrrr}
    \toprule
    \multirow{1}{*}{\textbf{Datasets}}      
    & \multicolumn{1}{c}{\textbf{Modality}}
    & \multicolumn{1}{c}{\textbf{Labels}} 
    & \multicolumn{1}{c}{\textbf{Images}}
    & \multicolumn{1}{c|}{\textbf{Masks}}
    & \multicolumn{1}{c}{\textbf{P-I}}
    & \multicolumn{1}{c}{\textbf{P-II}}
    & \multicolumn{1}{c}{\textbf{P-III}}
    & \multicolumn{1}{c}{\textbf{P-IV}}\\
    \midrule
            
    \multicolumn{1}{c|}{CT-ORG\cite{Rister2020CTORGAN}} 
    & \multicolumn{1}{c}{CT} 
    & \multicolumn{1}{c}{Kidney,Lung,brain,etc.} 
    & \multicolumn{1}{c}{30000}
    & \multicolumn{1}{c|}{76171} 
    & \multicolumn{1}{c}{\ding{52}}
    & \multicolumn{1}{c}{\ding{52}}
    & \multicolumn{1}{c}{\ding{56}}
    & \multicolumn{1}{c}{\ding{56}}
    \\
    
    \multicolumn{1}{c|}{BraTS2021\cite{baid2021rsna}} 
    & \multicolumn{1}{c}{MR(FLAIR,T1CE)} 
    & \multicolumn{1}{c}{Edema,Tumor} 
    & \multicolumn{1}{c}{59963}
    & \multicolumn{1}{c|}{117335} 
    & \multicolumn{1}{c}{\ding{52}}
    & \multicolumn{1}{c}{\ding{52}}
    & \multicolumn{1}{c}{\ding{56}}
    & \multicolumn{1}{c}{\ding{56}}
    \\

    \multicolumn{1}{c|}{BraTS2021\cite{baid2021rsna}} 
    & \multicolumn{1}{c}{MR(T1,T2)} 
    & \multicolumn{1}{c}{Edema} 
    & \multicolumn{1}{c}{58228}
    & \multicolumn{1}{c|}{58228} 
    & \multicolumn{1}{c}{\ding{56}}
    & \multicolumn{1}{c}{\ding{56}}
    & \multicolumn{1}{c}{\ding{52}}
    & \multicolumn{1}{c}{\ding{52}}
    \\
    
    \multicolumn{1}{c|}{EndoVis-2017-RIS\cite{allan20192017}} 
    & \multicolumn{1}{c}{Endoscopy} 
    & \multicolumn{1}{c}{Shaft, Wrist, Clasper} 
    & \multicolumn{1}{c}{2978}
    & \multicolumn{1}{c|}{8530} 
    & \multicolumn{1}{c}{\ding{52}}
    & \multicolumn{1}{c}{\ding{52}}
    & \multicolumn{1}{c}{\ding{56}}
    & \multicolumn{1}{c}{\ding{56}}
    \\

    \multicolumn{1}{c|}{endovis15\cite{du2018articulated}} 
    & \multicolumn{1}{c}{Endoscopy} 
    & \multicolumn{1}{c}{Polyp} 
    & \multicolumn{1}{c}{612}
    & \multicolumn{1}{c|}{612} 
    & \multicolumn{1}{c}{\ding{56}}
    & \multicolumn{1}{c}{\ding{56}}
    & \multicolumn{1}{c}{\ding{52}}
    & \multicolumn{1}{c}{\ding{52}}
    \\
    
    \multicolumn{1}{c|}{BrainTumour\cite{faa44e0a12da4c11aeee91cc3c8ac11e}} 
    & \multicolumn{1}{c}{MR(T1W,T2W)} 
    & \multicolumn{1}{c}{Edema,Tumor} 
    & \multicolumn{1}{c}{59951}
    & \multicolumn{1}{c|}{118607} 
    & \multicolumn{1}{c}{\ding{52}}
    & \multicolumn{1}{c}{\ding{52}}
    & \multicolumn{1}{c}{\ding{56}}
    & \multicolumn{1}{c}{\ding{56}}
    \\

    \multicolumn{1}{c|}{BrainTumour\cite{faa44e0a12da4c11aeee91cc3c8ac11e}} 
    & \multicolumn{1}{c}{MR-FLAIR} 
    & \multicolumn{1}{c}{Edema,Tumor} 
    & \multicolumn{1}{c}{29357}
    & \multicolumn{1}{c|}{29357} 
    & \multicolumn{1}{c}{\ding{56}}
    & \multicolumn{1}{c}{\ding{56}}
    & \multicolumn{1}{c}{\ding{52}}
    & \multicolumn{1}{c}{\ding{52}}
    \\
    
    \multicolumn{1}{c|}{ATM2022\cite{zhang2023multi}} 
    & \multicolumn{1}{c}{CT} 
    & \multicolumn{1}{c}{Respiratory Tract} 
    & \multicolumn{1}{c}{39227}
    & \multicolumn{1}{c|}{39227} 
    & \multicolumn{1}{c}{\ding{52}}
    & \multicolumn{1}{c}{\ding{52}}
    & \multicolumn{1}{c}{\ding{56}}
    & \multicolumn{1}{c}{\ding{56}}
    \\

    \multicolumn{1}{c|}{LNDb\cite{pedrosa2019lndb}} 
    & \multicolumn{1}{c}{CT} 
    & \multicolumn{1}{c}{Lung Nodule} 
    & \multicolumn{1}{c}{962}
    & \multicolumn{1}{c|}{962} 
    & \multicolumn{1}{c}{\ding{52}}
    & \multicolumn{1}{c}{\ding{52}}
    & \multicolumn{1}{c}{\ding{56}}
    & \multicolumn{1}{c}{\ding{56}}
    \\

    \multicolumn{1}{c|}{VerSe20\cite{sekuboyina2021verse}} 
    & \multicolumn{1}{c}{CT} 
    & \multicolumn{1}{c}{Vertebrae} 
    & \multicolumn{1}{c}{29997}
    & \multicolumn{1}{c|}{77775} 
    & \multicolumn{1}{c}{\ding{52}}
    & \multicolumn{1}{c}{\ding{52}}
    & \multicolumn{1}{c}{\ding{56}}
    & \multicolumn{1}{c}{\ding{56}}
    \\

    \multicolumn{1}{c|}{ISLES\_SISS\cite{maier2017isles}} 
    & \multicolumn{1}{c}{MR(DWI,T1)} 
    & \multicolumn{1}{c}{Ischemic Stroke} 
    & \multicolumn{1}{c}{6527}
    & \multicolumn{1}{c|}{6527} 
    & \multicolumn{1}{c}{\ding{52}}
    & \multicolumn{1}{c}{\ding{52}}
    & \multicolumn{1}{c}{\ding{56}}
    & \multicolumn{1}{c}{\ding{56}}
    \\

    \multicolumn{1}{c|}{ISLES\_SISS\cite{maier2017isles}} 
    & \multicolumn{1}{c}{MR(FLAIR,T2)} 
    & \multicolumn{1}{c}{Ischemic Stroke} 
    & \multicolumn{1}{c}{6538}
    & \multicolumn{1}{c|}{6538} 
    & \multicolumn{1}{c}{\ding{56}}
    & \multicolumn{1}{c}{\ding{56}}
    & \multicolumn{1}{c}{\ding{52}}
    & \multicolumn{1}{c}{\ding{52}}
    \\
    
    \multicolumn{1}{c|}{ISLES-SPES\cite{maier2017isles}} 
    & \multicolumn{1}{c}{MR(T2,DWI)} 
    & \multicolumn{1}{c}{Ischemic Stroke} 
    & \multicolumn{1}{c}{6541}
    & \multicolumn{1}{c|}{6541} 
    & \multicolumn{1}{c}{\ding{52}}
    & \multicolumn{1}{c}{\ding{52}}
    & \multicolumn{1}{c}{\ding{56}}
    & \multicolumn{1}{c}{\ding{56}}
    \\

    \multicolumn{1}{c|}{ISLES-SPES\cite{maier2017isles}} 
    & \multicolumn{1}{c}{MR(cbf,ttp,tmax,cbv,t1c)} 
    & \multicolumn{1}{c}{Ischemic Stroke} 
    & \multicolumn{1}{c}{16395}
    & \multicolumn{1}{c|}{16395} 
    & \multicolumn{1}{c}{\ding{56}}
    & \multicolumn{1}{c}{\ding{56}}
    & \multicolumn{1}{c}{\ding{52}}
    & \multicolumn{1}{c}{\ding{52}}
    \\
    
    \multicolumn{1}{c|}{ISLES2018\cite{maier2017isles}} 
    & \multicolumn{1}{c}{CT(TMAX,CBV)} 
    & \multicolumn{1}{c}{Ischemic Stroke} 
    & \multicolumn{1}{c}{565}
    & \multicolumn{1}{c|}{565} 
    & \multicolumn{1}{c}{\ding{52}}
    & \multicolumn{1}{c}{\ding{52}}
    & \multicolumn{1}{c}{\ding{56}}
    & \multicolumn{1}{c}{\ding{56}}
    \\

    \multicolumn{1}{c|}{ISLES2018\cite{maier2017isles}} 
    & \multicolumn{1}{c}{CT-MTT} 
    & \multicolumn{1}{c}{Ischemic Stroke} 
    & \multicolumn{1}{c}{283}
    & \multicolumn{1}{c|}{283} 
    & \multicolumn{1}{c}{\ding{56}}
    & \multicolumn{1}{c}{\ding{56}}
    & \multicolumn{1}{c}{\ding{52}}
    & \multicolumn{1}{c}{\ding{52}}
    \\
    
    \multicolumn{1}{c|}{ISLES2022\cite{article}} 
    & \multicolumn{1}{c}{MR-ADC} 
    & \multicolumn{1}{c}{Ischemic Stroke} 
    & \multicolumn{1}{c}{5919}
    & \multicolumn{1}{c|}{5919} 
    & \multicolumn{1}{c}{\ding{52}}
    & \multicolumn{1}{c}{\ding{52}}
    & \multicolumn{1}{c}{\ding{56}}
    & \multicolumn{1}{c}{\ding{56}}
    \\

    \multicolumn{1}{c|}{ISLES2022\cite{article}} 
    & \multicolumn{1}{c}{MR-DWI} 
    & \multicolumn{1}{c}{Ischemic Stroke} 
    & \multicolumn{1}{c}{5928}
    & \multicolumn{1}{c|}{5929} 
    & \multicolumn{1}{c}{\ding{56}}
    & \multicolumn{1}{c}{\ding{56}}
    & \multicolumn{1}{c}{\ding{56}}
    & \multicolumn{1}{c}{\ding{52}}
    \\
    
    \multicolumn{1}{c|}{Totalsegmentator-dataset\cite{TotalSegmentator}} 
    & \multicolumn{1}{c}{CT} 
    & \multicolumn{1}{c}{Spleen,Hip,Urinary Bladder,etc.} 
    & \multicolumn{1}{c}{140624}
    & \multicolumn{1}{c|}{1055734} 
    & \multicolumn{1}{c}{\ding{56}}
    & \multicolumn{1}{c}{\ding{56}}
    & \multicolumn{1}{c}{\ding{56}}
    & \multicolumn{1}{c}{\ding{52}}
    \\

    \multicolumn{1}{c|}{autoPET\cite{zhong2022autopet}} 
    & \multicolumn{1}{c}{PET} 
    & \multicolumn{1}{c}{Tumor} 
    & \multicolumn{1}{c}{6019}
    & \multicolumn{1}{c|}{6019} 
    & \multicolumn{1}{c}{\ding{52}}
    & \multicolumn{1}{c}{\ding{52}}
    & \multicolumn{1}{c}{\ding{56}}
    & \multicolumn{1}{c}{\ding{56}}
    \\

    \multicolumn{1}{c|}{VESSEL2012\cite{rudyanto2014comparing}} 
    & \multicolumn{1}{c}{CT} 
    & \multicolumn{1}{c}{Lung Vessel} 
    & \multicolumn{1}{c}{20405}
    & \multicolumn{1}{c|}{20405} 
    & \multicolumn{1}{c}{\ding{52}}
    & \multicolumn{1}{c}{\ding{52}}
    & \multicolumn{1}{c}{\ding{56}}
    & \multicolumn{1}{c}{\ding{56}}
    \\

    \multicolumn{1}{c|}{Brain-PTM\cite{avital2019neural}} 
    & \multicolumn{1}{c}{MR-T1} 
    & \multicolumn{1}{c}{Matter Tracts} 
    & \multicolumn{1}{c}{11402}
    & \multicolumn{1}{c|}{11402} 
    & \multicolumn{1}{c}{\ding{52}}
    & \multicolumn{1}{c}{\ding{52}}
    & \multicolumn{1}{c}{\ding{56}}
    & \multicolumn{1}{c}{\ding{56}}
    \\

    \multicolumn{1}{c|}{AMOS2022\cite{ji2022amos}} 
    & \multicolumn{1}{c}{MR} 
    & \multicolumn{1}{c}{Spleen,Duodenum,Esophagus,etc.} 
    & \multicolumn{1}{c}{6391}
    & \multicolumn{1}{c|}{28974} 
    & \multicolumn{1}{c}{\ding{52}}
    & \multicolumn{1}{c}{\ding{52}}
    & \multicolumn{1}{c}{\ding{56}}
    & \multicolumn{1}{c}{\ding{56}}
    \\

    \multicolumn{1}{c|}{AMOS2022\cite{ji2022amos}} 
    & \multicolumn{1}{c}{CT} 
    & \multicolumn{1}{c}{Spleen,Duodenum,Esophagus,etc.} 
    & \multicolumn{1}{c}{5985}
    & \multicolumn{1}{c|}{10169} 
    & \multicolumn{1}{c}{\ding{56}}
    & \multicolumn{1}{c}{\ding{56}}
    & \multicolumn{1}{c}{\ding{56}}
    & \multicolumn{1}{c}{\ding{52}}
    \\
    
     \multicolumn{1}{c|}{CTSpine1K-Full\cite{deng2021ctspine1k}} 
    & \multicolumn{1}{c}{CT} 
    & \multicolumn{1}{c}{Vertebrae} 
    & \multicolumn{1}{c}{29998}
    & \multicolumn{1}{c|}{82162} 
    & \multicolumn{1}{c}{\ding{52}}
    & \multicolumn{1}{c}{\ding{52}}
    & \multicolumn{1}{c}{\ding{56}}
    & \multicolumn{1}{c}{\ding{56}}
    \\

     \multicolumn{1}{c|}{AbdomenCT1K\cite{Ma-2021-AbdomenCT-1K}} 
    & \multicolumn{1}{c}{CT} 
    & \multicolumn{1}{c}{Pancreas,Kidney,Spleen,Liver} 
    & \multicolumn{1}{c}{40000}
    & \multicolumn{1}{c|}{86070} 
    & \multicolumn{1}{c}{\ding{52}}
    & \multicolumn{1}{c}{\ding{52}}
    & \multicolumn{1}{c}{\ding{56}}
    & \multicolumn{1}{c}{\ding{56}}
    \\

     \multicolumn{1}{c|}{CTPelvic1k\cite{liu2021deep}} 
    & \multicolumn{1}{c}{CT} 
    & \multicolumn{1}{c}{Hip,Sacrum,Lumbar Vertebra} 
    & \multicolumn{1}{c}{30000}
    & \multicolumn{1}{c|}{63882} 
    & \multicolumn{1}{c}{\ding{52}}
    & \multicolumn{1}{c}{\ding{52}}
    & \multicolumn{1}{c}{\ding{56}}
    & \multicolumn{1}{c}{\ding{56}}
    \\

     \multicolumn{1}{c|}{CrossMoDA22\cite{dorent2023crossmoda}} 
    & \multicolumn{1}{c}{MR-T1CE} 
    & \multicolumn{1}{c}{Vestibular Schwannoma} 
    & \multicolumn{1}{c}{1476}
    & \multicolumn{1}{c|}{1476} 
    & \multicolumn{1}{c}{\ding{52}}
    & \multicolumn{1}{c}{\ding{52}}
    & \multicolumn{1}{c}{\ding{56}}
    & \multicolumn{1}{c}{\ding{56}}
    \\

     \multicolumn{1}{c|}{Ultrasound-nerve\cite{ultrasound-nerve-segmentation}} 
    & \multicolumn{1}{c}{Ultrasound} 
    & \multicolumn{1}{c}{Ultrasound Nerve} 
    & \multicolumn{1}{c}{2320}
    & \multicolumn{1}{c|}{2320} 
    & \multicolumn{1}{c}{\ding{52}}
    & \multicolumn{1}{c}{\ding{52}}
    & \multicolumn{1}{c}{\ding{56}}
    & \multicolumn{1}{c}{\ding{56}}
    \\

     \multicolumn{1}{c|}{Isic2016-task1\cite{gutman2016skin}} 
    & \multicolumn{1}{c}{Dermoscopy} 
    & \multicolumn{1}{c}{Skin Lesion} 
    & \multicolumn{1}{c}{1279}
    & \multicolumn{1}{c|}{1279} 
    & \multicolumn{1}{c}{\ding{56}}
    & \multicolumn{1}{c}{\ding{56}}
    & \multicolumn{1}{c}{\ding{52}}
    & \multicolumn{1}{c}{\ding{52}}
    \\
    
     \multicolumn{1}{c|}{Isic2017-task1\cite{codella2018skin}} 
    & \multicolumn{1}{c}{Dermoscopy} 
    & \multicolumn{1}{c}{Skin Lesion} 
    & \multicolumn{1}{c}{2746}
    & \multicolumn{1}{c|}{2746} 
    & \multicolumn{1}{c}{\ding{52}}
    & \multicolumn{1}{c}{\ding{52}}
    & \multicolumn{1}{c}{\ding{56}}
    & \multicolumn{1}{c}{\ding{56}}
    \\

     \multicolumn{1}{c|}{Isic2018-task1\cite{codella2019skin}} 
    & \multicolumn{1}{c}{Dermoscopy} 
    & \multicolumn{1}{c}{Skin Lesion} 
    & \multicolumn{1}{c}{2689}
    & \multicolumn{1}{c|}{2689} 
    & \multicolumn{1}{c}{\ding{52}}
    & \multicolumn{1}{c}{\ding{52}}
    & \multicolumn{1}{c}{\ding{56}}
    & \multicolumn{1}{c}{\ding{56}}
    \\

     \multicolumn{1}{c|}{LongitudinalMultiple\cite{carass2017longitudinal}} 
    & \multicolumn{1}{c}{MR-T2} 
    & \multicolumn{1}{c}{Multiple Sclerosis Lesion} 
    & \multicolumn{1}{c}{2025}
    & \multicolumn{1}{c|}{2025} 
    & \multicolumn{1}{c}{\ding{52}}
    & \multicolumn{1}{c}{\ding{52}}
    & \multicolumn{1}{c}{\ding{56}}
    & \multicolumn{1}{c}{\ding{56}}
    \\

     \multicolumn{1}{c|}{LongitudinalMultiple\cite{carass2017longitudinal}} 
    & \multicolumn{1}{c}{MR-FLAIR} 
    & \multicolumn{1}{c}{Multiple Sclerosis Lesion} 
    & \multicolumn{1}{c}{976}
    & \multicolumn{1}{c|}{976} 
    & \multicolumn{1}{c}{\ding{56}}
    & \multicolumn{1}{c}{\ding{56}}
    & \multicolumn{1}{c}{\ding{52}}
    & \multicolumn{1}{c}{\ding{52}}
    \\
    
     \multicolumn{1}{c|}{mnms2\cite{mazher2021multi}} 
    & \multicolumn{1}{c}{MR} 
    & \multicolumn{1}{c}{Ventricle,Ventricular Myocardium,etc.} 
    & \multicolumn{1}{c}{2477}
    & \multicolumn{1}{c|}{6779} 
    & \multicolumn{1}{c}{\ding{52}}
    & \multicolumn{1}{c}{\ding{52}}
    & \multicolumn{1}{c}{\ding{56}}
    & \multicolumn{1}{c}{\ding{56}}
    \\

     \multicolumn{1}{c|}{COVID19CTscans\cite{Rahimzadeh2020.06.08.20121541}} 
    & \multicolumn{1}{c}{CT} 
    & \multicolumn{1}{c}{Lung,Lung Infectiions} 
    & \multicolumn{1}{c}{6719}
    & \multicolumn{1}{c|}{14255} 
    & \multicolumn{1}{c}{\ding{52}}
    & \multicolumn{1}{c}{\ding{52}}
    & \multicolumn{1}{c}{\ding{56}}
    & \multicolumn{1}{c}{\ding{56}}
    \\

     \multicolumn{1}{c|}{ASC18\cite{cheng2023sam}} 
    & \multicolumn{1}{c}{MR-LGE} 
    & \multicolumn{1}{c}{Left Atrium} 
    & \multicolumn{1}{c}{8210}
    & \multicolumn{1}{c|}{8210} 
    & \multicolumn{1}{c}{\ding{52}}
    & \multicolumn{1}{c}{\ding{52}}
    & \multicolumn{1}{c}{\ding{56}}
    & \multicolumn{1}{c}{\ding{56}}
    \\

     \multicolumn{1}{c|}{cvc-clinicdb\cite{cvc}} 
    & \multicolumn{1}{c}{Endoscopy} 
    & \multicolumn{1}{c}{Polyp} 
    & \multicolumn{1}{c}{612}
    & \multicolumn{1}{c|}{612} 
    & \multicolumn{1}{c}{\ding{52}}
    & \multicolumn{1}{c}{\ding{52}}
    & \multicolumn{1}{c}{\ding{56}}
    & \multicolumn{1}{c}{\ding{56}}
    \\

     \multicolumn{1}{c|}{hvsmr-2016\cite{hvsmr}} 
    & \multicolumn{1}{c}{MR} 
    & \multicolumn{1}{c}{Heart Blood Pool,Myocardium,etc.} 
    & \multicolumn{1}{c}{1979}
    & \multicolumn{1}{c|}{5224} 
    & \multicolumn{1}{c}{\ding{52}}
    & \multicolumn{1}{c}{\ding{52}}
    & \multicolumn{1}{c}{\ding{56}}
    & \multicolumn{1}{c}{\ding{56}}
    \\

     \multicolumn{1}{c|}{PALM\cite{55pk-8z03-19}} 
    & \multicolumn{1}{c}{Fundus-Photography} 
    & \multicolumn{1}{c}{Optic Disc} 
    & \multicolumn{1}{c}{1144}
    & \multicolumn{1}{c|}{1144} 
    & \multicolumn{1}{c}{\ding{52}}
    & \multicolumn{1}{c}{\ding{52}}
    & \multicolumn{1}{c}{\ding{56}}
    & \multicolumn{1}{c}{\ding{56}}
    \\

     \multicolumn{1}{c|}{COVID-19-20\cite{roth2022rapid}} 
    & \multicolumn{1}{c}{CT} 
    & \multicolumn{1}{c}{COVID} 
    & \multicolumn{1}{c}{4794}
    & \multicolumn{1}{c|}{4794} 
    & \multicolumn{1}{c}{\ding{52}}
    & \multicolumn{1}{c}{\ding{52}}
    & \multicolumn{1}{c}{\ding{56}}
    & \multicolumn{1}{c}{\ding{56}}
    \\

     \multicolumn{1}{c|}{Prostate-MRI\cite{10.1007/978-3-030-59713-9_46}} 
    & \multicolumn{1}{c}{MR-T2W} 
    & \multicolumn{1}{c}{Prostate} 
    & \multicolumn{1}{c}{1861}
    & \multicolumn{1}{c|}{1861} 
    & \multicolumn{1}{c}{\ding{52}}
    & \multicolumn{1}{c}{\ding{52}}
    & \multicolumn{1}{c}{\ding{56}}
    & \multicolumn{1}{c}{\ding{56}}
    \\

     \multicolumn{1}{c|}{MSD-Lung\cite{simpson2019largeannotatedmedicalimage}} 
    & \multicolumn{1}{c}{CT} 
    & \multicolumn{1}{c}{Lung Cancer} 
    & \multicolumn{1}{c}{2313}
    & \multicolumn{1}{c|}{2313} 
    & \multicolumn{1}{c}{\ding{52}}
    & \multicolumn{1}{c}{\ding{52}}
    & \multicolumn{1}{c}{\ding{56}}
    & \multicolumn{1}{c}{\ding{56}}
    \\

     \multicolumn{1}{c|}{MSD-Spleen\cite{simpson2019largeannotatedmedicalimage}} 
    & \multicolumn{1}{c}{CT} 
    & \multicolumn{1}{c}{Spleen} 
    & \multicolumn{1}{c}{1004}
    & \multicolumn{1}{c|}{1004} 
    & \multicolumn{1}{c}{\ding{56}}
    & \multicolumn{1}{c}{\ding{56}}
    & \multicolumn{1}{c}{\ding{52}}
    & \multicolumn{1}{c}{\ding{52}}
    \\

     \multicolumn{1}{c|}{MSD-Heart\cite{simpson2019largeannotatedmedicalimage}} 
    & \multicolumn{1}{c}{MR} 
    & \multicolumn{1}{c}{Left Atrium} 
    & \multicolumn{1}{c}{1081}
    & \multicolumn{1}{c|}{1081} 
    & \multicolumn{1}{c}{\ding{56}}
    & \multicolumn{1}{c}{\ding{56}}
    & \multicolumn{1}{c}{\ding{52}}
    & \multicolumn{1}{c}{\ding{52}}
    \\

     \multicolumn{1}{c|}{MSD-Prostate\cite{simpson2019largeannotatedmedicalimage}} 
    & \multicolumn{1}{c}{MR-ADC} 
    & \multicolumn{1}{c}{Prostate Zone} 
    & \multicolumn{1}{c}{365}
    & \multicolumn{1}{c|}{497} 
    & \multicolumn{1}{c}{\ding{56}}
    & \multicolumn{1}{c}{\ding{56}}
    & \multicolumn{1}{c}{\ding{52}}
    & \multicolumn{1}{c}{\ding{52}}
    \\
    
    \multicolumn{1}{c|}{Instance22\cite{siddiquee2022automatedsegmentationintracranialhemorrhages}} 
    & \multicolumn{1}{c}{CT} 
    & \multicolumn{1}{c}{Intracranial Hemorrhage} 
    & \multicolumn{1}{c}{698}
    & \multicolumn{1}{c|}{698} 
    & \multicolumn{1}{c}{\ding{52}}
    & \multicolumn{1}{c}{\ding{52}}
    & \multicolumn{1}{c}{\ding{56}}
    & \multicolumn{1}{c}{\ding{56}}
    \\

    \multicolumn{1}{c|}{picai-semi\cite{saha2024artificial}} 
    & \multicolumn{1}{c}{MR-HBV} 
    & \multicolumn{1}{c}{Prostate Cancer} 
    & \multicolumn{1}{c}{414}
    & \multicolumn{1}{c|}{414} 
    & \multicolumn{1}{c}{\ding{52}}
    & \multicolumn{1}{c}{\ding{52}}
    & \multicolumn{1}{c}{\ding{56}}
    & \multicolumn{1}{c}{\ding{56}}
    \\

    \multicolumn{1}{c|}{picai-baseline\cite{saha2024artificial}} 
    & \multicolumn{1}{c}{MR-HBV} 
    & \multicolumn{1}{c}{Prostate Cancer} 
    & \multicolumn{1}{c}{314}
    & \multicolumn{1}{c|}{314} 
    & \multicolumn{1}{c}{\ding{56}}
    & \multicolumn{1}{c}{\ding{56}}
    & \multicolumn{1}{c}{\ding{52}}
    & \multicolumn{1}{c}{\ding{52}}
    \\
    
    \multicolumn{1}{c|}{Pulmonary-Chest\cite{stefan_jaeger__2014}} 
    & \multicolumn{1}{c}{X-Ray} 
    & \multicolumn{1}{c}{Left Lung,Right Lung} 
    & \multicolumn{1}{c}{36}
    & \multicolumn{1}{c|}{72} 
    & \multicolumn{1}{c}{\ding{52}}
    & \multicolumn{1}{c}{\ding{52}}
    & \multicolumn{1}{c}{\ding{56}}
    & \multicolumn{1}{c}{\ding{56}}
    \\

    \multicolumn{1}{c|}{Parse22\cite{luo2024efficientautomaticsegmentationmultilevel}} 
    & \multicolumn{1}{c}{CT} 
    & \multicolumn{1}{c}{Pulmonary Artery} 
    & \multicolumn{1}{c}{34684}
    & \multicolumn{1}{c|}{34684} 
    & \multicolumn{1}{c}{\ding{56}}
    & \multicolumn{1}{c}{\ding{56}}
    & \multicolumn{1}{c}{\ding{56}}
    & \multicolumn{1}{c}{\ding{52}}
    \\

    \multicolumn{1}{c|}{KiTS\cite{heller2020state}} 
    & \multicolumn{1}{c}{CT} 
    & \multicolumn{1}{c}{Kidney Tumor,Kidney} 
    & \multicolumn{1}{c}{33112}
    & \multicolumn{1}{c|}{40874} 
    & \multicolumn{1}{c}{\ding{56}}
    & \multicolumn{1}{c}{\ding{56}}
    & \multicolumn{1}{c}{\ding{52}}
    & \multicolumn{1}{c}{\ding{52}}
    \\
    
    \multicolumn{1}{c|}{ACDC\cite{sakaridis2024acdcadverseconditionsdataset}} 
    & \multicolumn{1}{c}{MR} 
    & \multicolumn{1}{c}{Ventricle,myocardium} 
    & \multicolumn{1}{c}{1835}
    & \multicolumn{1}{c|}{3554} 
    & \multicolumn{1}{c}{\ding{56}}
    & \multicolumn{1}{c}{\ding{56}}
    & \multicolumn{1}{c}{\ding{52}}
    & \multicolumn{1}{c}{\ding{52}}
    \\

    \multicolumn{1}{c|}{LUNA16\cite{setio2017validation}} 
    & \multicolumn{1}{c}{CT} 
    & \multicolumn{1}{c}{Lung} 
    & \multicolumn{1}{c}{30000}
    & \multicolumn{1}{c|}{30000} 
    & \multicolumn{1}{c}{\ding{56}}
    & \multicolumn{1}{c}{\ding{56}}
    & \multicolumn{1}{c}{\ding{56}}
    & \multicolumn{1}{c}{\ding{52}}
    \\

    \multicolumn{1}{c|}{MMWHS\cite{gao2023bayeseg}} 
    & \multicolumn{1}{c}{CT} 
    & \multicolumn{1}{c}{Pulmonary Artery,Atrium Blood Cavity,etc.} 
    & \multicolumn{1}{c}{2199}
    & \multicolumn{1}{c|}{3368} 
    & \multicolumn{1}{c}{\ding{56}}
    & \multicolumn{1}{c}{\ding{56}}
    & \multicolumn{1}{c}{\ding{52}}
    & \multicolumn{1}{c}{\ding{52}}
    \\

    \multicolumn{1}{c|}{FLARE21\cite{MedIA-FLARE21}} 
    & \multicolumn{1}{c}{CT} 
    & \multicolumn{1}{c}{Pancreas,Kidney,Liver,Spleen} 
    & \multicolumn{1}{c}{20000}
    & \multicolumn{1}{c|}{34454} 
    & \multicolumn{1}{c}{\ding{56}}
    & \multicolumn{1}{c}{\ding{56}}
    & \multicolumn{1}{c}{\ding{56}}
    & \multicolumn{1}{c}{\ding{52}}
    \\

    \multicolumn{1}{c|}{CAD-PE\cite{9bw7-6823-19}} 
    & \multicolumn{1}{c}{CT} 
    & \multicolumn{1}{c}{Pulmonary Embolism} 
    & \multicolumn{1}{c}{6667}
    & \multicolumn{1}{c|}{6667} 
    & \multicolumn{1}{c}{\ding{56}}
    & \multicolumn{1}{c}{\ding{56}}
    & \multicolumn{1}{c}{\ding{56}}
    & \multicolumn{1}{c}{\ding{52}}
    \\

    \multicolumn{1}{c|}{Chest-Image-Pneum\cite{kermany2018labeled}} 
    & \multicolumn{1}{c}{X-Ray} 
    & \multicolumn{1}{c}{Pneumothorax} 
    & \multicolumn{1}{c}{2426}
    & \multicolumn{1}{c|}{2426} 
    & \multicolumn{1}{c}{\ding{56}}
    & \multicolumn{1}{c}{\ding{56}}
    & \multicolumn{1}{c}{\ding{52}}
    & \multicolumn{1}{c}{\ding{52}}
    \\

    \multicolumn{1}{c|}{WORD\cite{luo2022word}} 
    & \multicolumn{1}{c}{CT} 
    & \multicolumn{1}{c}{Femur Head,Stomach,Spleen,Duodenum,etc.} 
    & \multicolumn{1}{c}{16571}
    & \multicolumn{1}{c|}{28742} 
    & \multicolumn{1}{c}{\ding{56}}
    & \multicolumn{1}{c}{\ding{56}}
    & \multicolumn{1}{c}{\ding{52}}
    & \multicolumn{1}{c}{\ding{52}}
    \\

    \multicolumn{1}{c|}{PROMISE12\cite{LITJENS2014359}} 
    & \multicolumn{1}{c}{MR} 
    & \multicolumn{1}{c}{Prostate} 
    & \multicolumn{1}{c}{776}
    & \multicolumn{1}{c|}{776} 
    & \multicolumn{1}{c}{\ding{56}}
    & \multicolumn{1}{c}{\ding{56}}
    & \multicolumn{1}{c}{\ding{52}}
    & \multicolumn{1}{c}{\ding{52}}
    \\

    \multicolumn{1}{c|}{gamma\cite{wu2023gamma}} 
    & \multicolumn{1}{c}{Fundus-Photography} 
    & \multicolumn{1}{c}{Optic Disc,Optic Cup} 
    & \multicolumn{1}{c}{180}
    & \multicolumn{1}{c|}{180} 
    & \multicolumn{1}{c}{\ding{56}}
    & \multicolumn{1}{c}{\ding{56}}
    & \multicolumn{1}{c}{\ding{52}}
    & \multicolumn{1}{c}{\ding{52}}
    \\

    \multicolumn{1}{c|}{cranium\cite{cheng2023sam}} 
    & \multicolumn{1}{c}{CT} 
    & \multicolumn{1}{c}{Intracranial Hemorrhage} 
    & \multicolumn{1}{c}{210}
    & \multicolumn{1}{c|}{210} 
    & \multicolumn{1}{c}{\ding{56}}
    & \multicolumn{1}{c}{\ding{56}}
    & \multicolumn{1}{c}{\ding{52}}
    & \multicolumn{1}{c}{\ding{52}}
    \\
    
    \bottomrule
    \end{tabular}
    \label{tab:datasource}
    \vspace{-0.3cm}
\end{table*}

\end{document}